\documentclass{ametsocV6.1}



\usepackage{amssymb}
\usepackage{array}
\usepackage{makecell}
\usepackage{lscape}

\title{Generative ensemble deep learning severe weather prediction from a deterministic convection-allowing model}

\authors{
    Yingkai Sha,\aff{a}\correspondingauthor{ksha@ucar.edu}
    Ryan A. Sobash,\aff{a}
    David John Gagne II,\aff{a}}
    
\affiliation{
    \aff{a}{NSF National Center for Atmospheric Research, Boulder, Colorado, USA}}

\abstract{An ensemble post-processing method is developed for the probabilistic prediction of severe weather (tornadoes, hail, and wind gusts) over the conterminous United States (CONUS). The method combines conditional generative adversarial networks (CGANs), a type of deep generative model, with a convolutional neural network (CNN) to post-process convection-allowing model (CAM) forecasts. The CGANs are designed to create synthetic ensemble members from deterministic CAM forecasts, and their outputs are processed by the CNN to estimate the probability of severe weather. The method is tested using High-Resolution Rapid Refresh (HRRR) 1--24 hr forecasts as inputs and Storm Prediction Center (SPC) severe weather reports as targets. The method produced skillful predictions with up to 20\% Brier Skill Score (BSS) increases compared to other neural-network-based reference methods using a testing dataset of HRRR forecasts in 2021. For the evaluation of uncertainty quantification, the method is overconfident but produces meaningful ensemble spreads that can distinguish good and bad forecasts. The quality of CGAN outputs is also evaluated. Results show that the CGAN outputs behave similarly to a numerical ensemble; they preserved the inter-variable correlations and the contribution of influential predictors as in the original HRRR forecasts. This work provides a novel approach to post-process CAM output using neural networks that can be applied to severe weather prediction.
}

\begin{document}

\maketitle

%
\statement
We use a new machine learning (ML) technique to generate probabilistic forecasts of convective weather hazards, such as tornadoes and hail storms, with the output from high-resolution numerical weather model forecasts. The new ML system generates an ensemble of synthetic forecast fields from a single forecast, which are then used to train ML models for convective hazard prediction. Using this ML-generated ensemble for training leads to improvements of 10--20\% in severe weather forecast skills compared to using other ML algorithms that use only output from the single forecast. This work is unique in that it explores the use of ML methods for producing synthetic forecasts of convective storm events and using these to train ML systems for high-impact convective weather prediction.


\section{Introduction}\label{sec1}

Convection-allowing models (CAMs) are numerical forecasting tools that have been applied routinely to provide guidance on fine spatial scales \citep{benjamin2016,dowell2022,james2022}. CAM forecasts can partially resolve storm-scale structures, and thus, they are particularly useful in diagnosing severe weather events, such as tornadoes, hail, and wind gusts \citep{kain2006,smith2012,roberts2019}.

Many post-processing studies have been conducted to extract severe-weather-based information from CAMs. Initially, severe weather probabilities were primarily derived from heuristic methods [e.g. \citet{theis2005} and \citet{roberts2005} for extreme precipitation, \citet{sobash2011,sobash2016} for severe thunderstorms, \citet{gallo2016,gallo2018} for tornadoes]. These methods convert CAM diagnostics into binary or probability values based on thresholds and use spatial smoothing operations to produce more skillful and visually appealing results. More recently, machine learning (ML) models have also been applied to the post-processing of CAM forecasts. For example, \citet{hill2020,hill2021,loken2020,loken2022} developed and compared differently designed random forests on severe weather or extreme rainfall predictions, \citet{gagne2017} applied random forests to produce probabilistic hail forecasts from CAM ensembles, \citet{lagerquist2020} applied Convolutional Neural Networks (CNNs) for next-hour tornado prediction, and \citet{sobash2020} trained Multilayer Perceptrons (MLPs) with improved feature engineering to produce point-based severe weather probabilities

Validating the probabilistic severe weather predictions and exploring the advances of deep generative models are the main motivations of this work. Ensemble predictions have clear benefits. The best guess of an ensemble is typically more skillful than deterministic predictions. The ensemble spread measures the predictive uncertainties; it answers the question of ``how much can we trust the predicted results?''. Given the difficulty and social impact of severe weather post-processing, organizing it within an ensemble prediction workflow is essential. Many deterministic, point-based machine learning methods can make ensemble predictions if they have an ensemble of CAM runs as inputs. However, maintaining and pre-processing a large ensemble of CAMs may introduce technical challenges such as data transmission and storage. The development of deep generative models provides an alternative. In this research, deep generative models are trained to create an ensemble of synthetic forecasts from a deterministic CAM run. This synthetic ensemble will provide convective-scale information and support the estimation of severe weather. Deep generative models are neural networks that can approximate probability distributions, and notably, they can create synthetic outputs from the data distributions they have learned \citep{creswell2018,goodfellow2014,ruthotto2021}. Deep generative models have achieved success in downscaling [e.g., \citet{leinonen2020} for satellite imagery, \citet{miralles2022} for surface wind], nowcasting [e.g., \citet{ravuri2021}, \citet{luo2022}, and \citet{tian2019} for precipitation nowcasting, \citet{gong2023} for radar reflectivity], and data-driven ensemble weather forecasting [\citet{zhong2023fuxi} for FuXi-Extreme, \citet{price2023gencast} for GenCast] problems, indicating that they can be applied to generate high-resolution meteorological fields.

In this research, a novel ML-based CAM post-processing method is introduced that combines a CNN-based prediction model and a deep generative model that creates synthetic CAM forecasts as input predictors for severe weather post-processing. The generative model can effectively generate an ensemble of severe weather predictions from a deterministic CAM run, potentially improving forecast skills and uncertainty estimates. The post-processing system uses CAM output on the native model grid, eliminating the need for upscaling that has been used in prior work \citep[e.g.][]{loken2020,sobash2020}. The system is trained to output probabilities of severe weather at grid points within the CAM domain using observations of severe weather events.

The deep generative model and the severe weather prediction model of this research are developed based on CNNs. CNNs are deep learning models that have hierarchically assigned spatial operations, and they are specialized in gridded, pattern-based learning problems \citep{aloysius2017,gu2018,goodfellow2016}. For post-processing studies, \citet{lerch2022} applied CNN-based autoencoders for the bias correction of 2-m air temperature and 850 hPa wind. \citet{li2022} applied a CNN regression model to the point-based bias correction of precipitation forecast. \citet{sha2022} applied an encoder-decoder CNN to improve precipitation analog ensembles in a complex terrain environment. The success of existing studies indicates that, if well-designed, CNNs can extract useful information from numerical forecasts. Yet, no prior work has applied CNNs for severe weather prediction at lead times beyond 1--2 hours, potentially due to the presence of larger model errors at longer lead times.

The proposed deep generative models and CNNs are trained and tested in the Conterminous United States (CONUS) using High-Resolution Rapid Refresh (HRRR) as inputs and the Storm Prediction Center (SPC) severe weather reports as targets. The following research questions are addressed: (1) How well can CNNs predict severe weather probabilities from high-resolution CAM fields? (2) Can we derive ensemble severe weather predictions from deterministic CAM forecasts? (3) What can deep generative models contribute within the context of severe weather predictions? By answering these, the authors aim to develop a post-processing system that provides skillful severe weather forecasts in CONUS. Broadly, the authors also wish to introduce deep generative models to CAM-related studies and inspire more creative works in the future.

\section{Research domain and data}\label{sec2}

\subsection{Region of interest}\label{sec21}

\begin{figure}[t]
 \noindent\includegraphics[width=\textwidth]{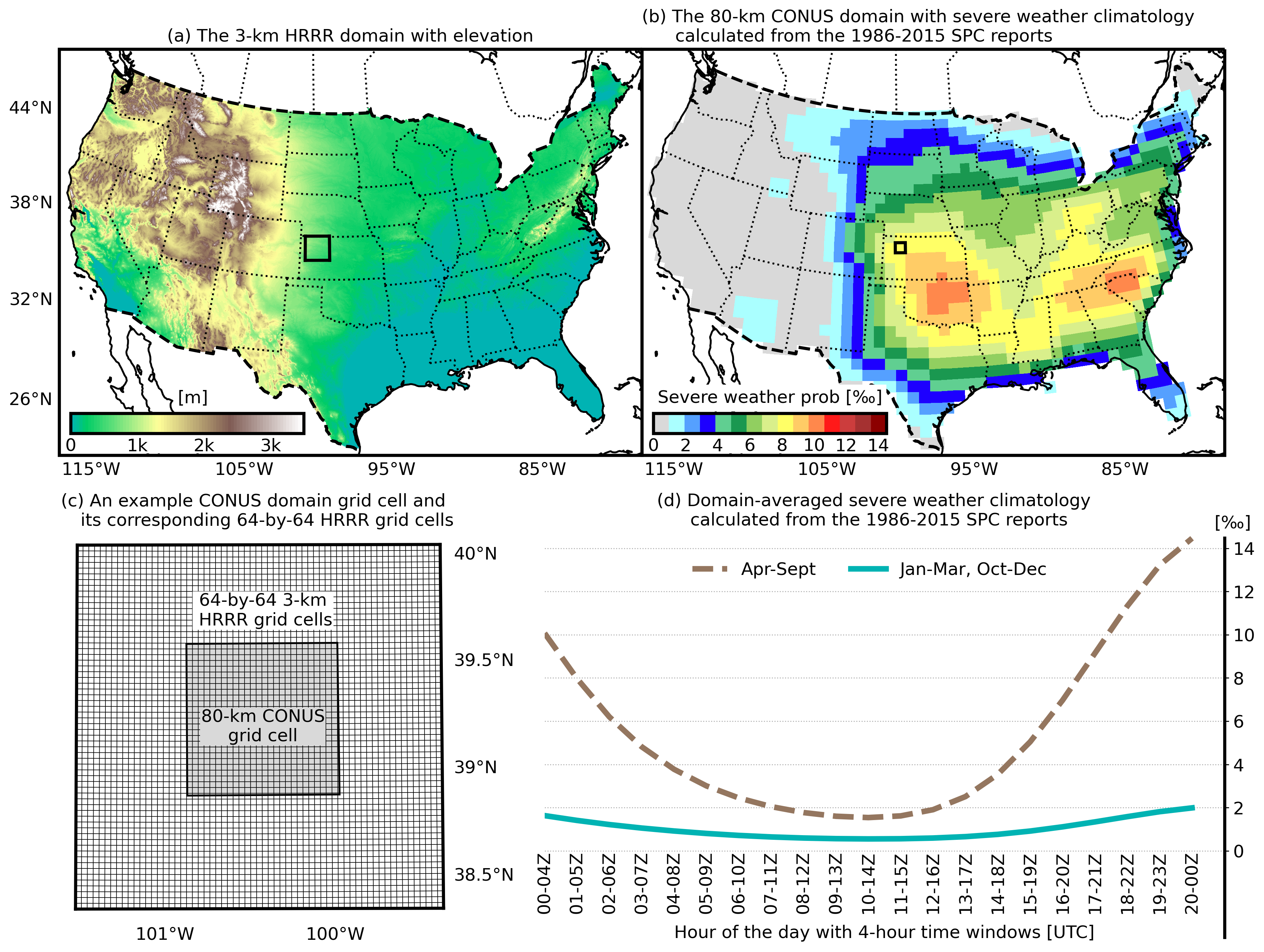}\\
 \caption{The 3-km grid spacing HRRR domain with shaded elevation. (b) The 80-km grid spacing CONUS domain with grid-point-wise severe weather climatological probabilities. (c) An example of a CONUS domain grid cell and its surrounding 64-by-64 HRRR cells. The locations and boundaries of grid cells are shown in (a) and (b) with black sold lines. (d) The domain-averaged severe weather climatology in Apr-Sept (dash line), and Jan-Mar, Oct-Dec (solid line) as functions of hour of the day with 4-hour time window.}\label{fig1}
\end{figure}

For this work, we focus on severe weather events occurring within the CONUS, which has a diverse range of climatological and geographical conditions that influence the spatial distribution of severe weather (Fig.~\ref{fig1}). Overall, the east side of the Rocky Mountains exhibits more frequent thunderstorms because of the impact of warm moist air from the Gulf of Mexico. These environmental conditions favor the occurrence of tornadoes, hail, and intense convective wind gusts, resulting in higher climatological probabilities. Regionally, the Central Plains have the highest severe weather probabilities due to a relatively larger number of tornado and hail reports compared to other regions. Severe weather probabilities are also high in the Southeast and lower-mid Atlantic due to an abundance of wind gust reports. Diurnal cycles and seasonal variations can also be found in severe weather climatologies (Fig.~\ref{fig1}.d). More than 40\% of the severe weather cases are reported during the local time of 12 UTC to 00 UTC and the warm season of April to September.

\subsection{Forecast data}\label{sec22}

\begin{landscape}
\begin{table}[t]
\caption{List of 15 predictors with abbreviations, variable types, and data normalization methods.}\label{tab1}
\centering
\begin{tabular}{lccc}
\hline\hline
Name & Abbreviation & Type & Data normalization \\
\hline
Latitude  & - & Static & [0, 1] feature scaling\\
Longitude & - & Static & [0, 1] feature scaling\\
Elevation & - & Static & [0, 1] feature scaling\\
Composite radar reflectivity & CREF & Explicit & Log transformation \\
Hourly maximum 0-2 km Above Ground Level (AGL) updraft helicity & 0-2 km UH & Explicit & Log transformation \\
Hourly maximum 2-5 km AGL updraft helicity  & 2-5 km UH & Explicit & Log transformation \\
Hourly accumulated precipitation & APCP & Explicit & Log transformation \\
Hourly maximum 10-m AGL wind speed & 10-m SPD & Explicit& Log transformation \\
Hourly maximum column graupel mass & GRPL & Explicit & Log transformation \\
Mean sea level pressure & MSLP & Environment & Standardization \\
2-m air temperature & 2-m Temp & Environment & Standardization \\
2-m dewpoint temperature & 2-m Dewpoint   & Environment & Standardization \\
Surface-based convective available potential energy & CAPE & Environment & Log transformation \\
Surface-based convective inhibition (magnitude) & CIN & Environment & Log transformation \\
0-1 km AGL storm-relative helicity & 0-1 km SRH & Environment & Standardization \\
0-3 km AGL storm-relative helicity & 0-3 km SRH & Environment & Standardization \\
0-6 km AGL u component wind shear & 0-6 km U shear & Environment & Standardization \\
0-6 km AGL u component wind shear & 0-6 km V shear & Environment & Standardization \\
\hline
\end{tabular}
\end{table}
\end{landscape}


We used both HRRR version 3 and 4 \citep{dowell2022}. The HRRR is an operational, real-time CAM based on WRF-ARW [Weather Research and Forecasting model, WRF; Advanced Research WRF, ARW; \citet{skamarock2019}]. HRRR has 3 km horizontal grid spacing and 50 vertical levels; it takes Rapid Refresh (RAP; \citet{benjamin2016}) as initial and boundary conditions, producing hourly, deterministic forecasts over CONUS. Here, the 0000 UTC initializations were used, with HRRRv3 covering 12 July 2018 to 2 December 2020, and HRRRv4 afterwards. In addition, experimental HRRRv4 initialized from 1 October 2019 to 2 December 2020 was also used.

Geographical inputs of latitude, longitude, and elevation, together with 15 HRRR diagnostics, were applied as predictors (Table.~\ref{tab1}). Storm-scale explicit predictors are surrogates for the potential occurrence of convective hazards, whereas environmental predictors are used to identify atmospheric conditions favorable for the occurrence of severe convection. Hereafter, the abbreviations of these predictors (Table~\ref{tab1}), will be used.

Predictions were generated on the 80-km grid across the CONUS from \citet{sobash2020}. The 15 HRRR predictors were pre-processed in two steps. First, for each 80-km grid cell, its center location was projected to the HRRR domain and subsets of 64-by-64-grid-cell HRRR predictors around the center location were extracted from the model grid (see Fig.~\ref{fig1}.c for an example). Then, these predictor subsets were normalized individually, using either logarithm transformation (CREF, 0-2 km UH, 2-5 km UH, APCP, 10-m SPD, GRPL, CAPE, CIN) or standardization (MSLP, 2-m Temp, 2-m Dewpoint, 0-1 km SRH, 0-3 km SRH, 0-6 km U shear, 0-6 km V shear). Geographical inputs are normalized with $[0, 1]$ feature scaling over the entire CONUS domain.

We used both HRRRv3 and HRRRv4 for training, even though changes in the dynamics and physics leads to a different climatology for some of the diagnostics \citep{dowell2022}. Notably, explicit predictors, such as 2-5 km UH, exhibit much lower magnitudes in HRRRv3 compared to that of the HRRRv4, This is likely because the two HRRR versions have different upper limits on the heating rate produced by the cloud microphysics scheme \citep[e.g.][]{wicker2020}. Here, we have compared the data distributions of all the normalized predictors, and we can confirm that after logarithm transformation, the magnitude differences of explicit predictors have been largely reduced. Thus, we think both HRRR versions can be applied to the training of post-processing models. In addition, as it will be clarified in the methods section, the verification of this research is based on HRRRv4 forecasts only. The overfitting of HRRRv3 forecasts, if any, will not be rewarded by verification results.

\subsection{Observations}\label{sec23}
Severe weather reports collected by the SPC at the National Oceanic and Atmospheric Administration (hereafter, ``SPC reports'') were used as the observational target. SPC reports provide the location and time for tornadoes, hail, and convective wind gusts that have sufficient intensities and societal impact. Such information is placed on the gridded 80-km CONUS domain by matching the closest grid cell to the starting location of each report. For temporal dimensions, SPC reports were aggregated hourly with a 4-hour time window. That said, given a severe weather report on hour $t$; it will be mapped to the 80-km CONUS domain four times, on $t-2$, $t-1$, $t$, and $t+1$. This choice is based on the operational demand for 4-hour severe weather outlooks \citep[e.g.][]{krocak2020}. 

Gridded SPC reports were used for the training and verification of the post-processing models, as well as the estimation of climatology references (Fig.~\ref{fig1}.b and d). We prefer the gridded SPC reports over the original ones because the spatiotemporal re-gridding creates a ``tolerance'' for the regional and population bias that SPC reports may potentially have \citep[e.g.][]{doswell1999}. Hereafter, when ``SPC reports'' is mentioned, it means the re-gridded version on the 80-km CONUS domain.

\section{Methods}\label{sec3}

\begin{figure}[t]
 \noindent\includegraphics[width=\textwidth]{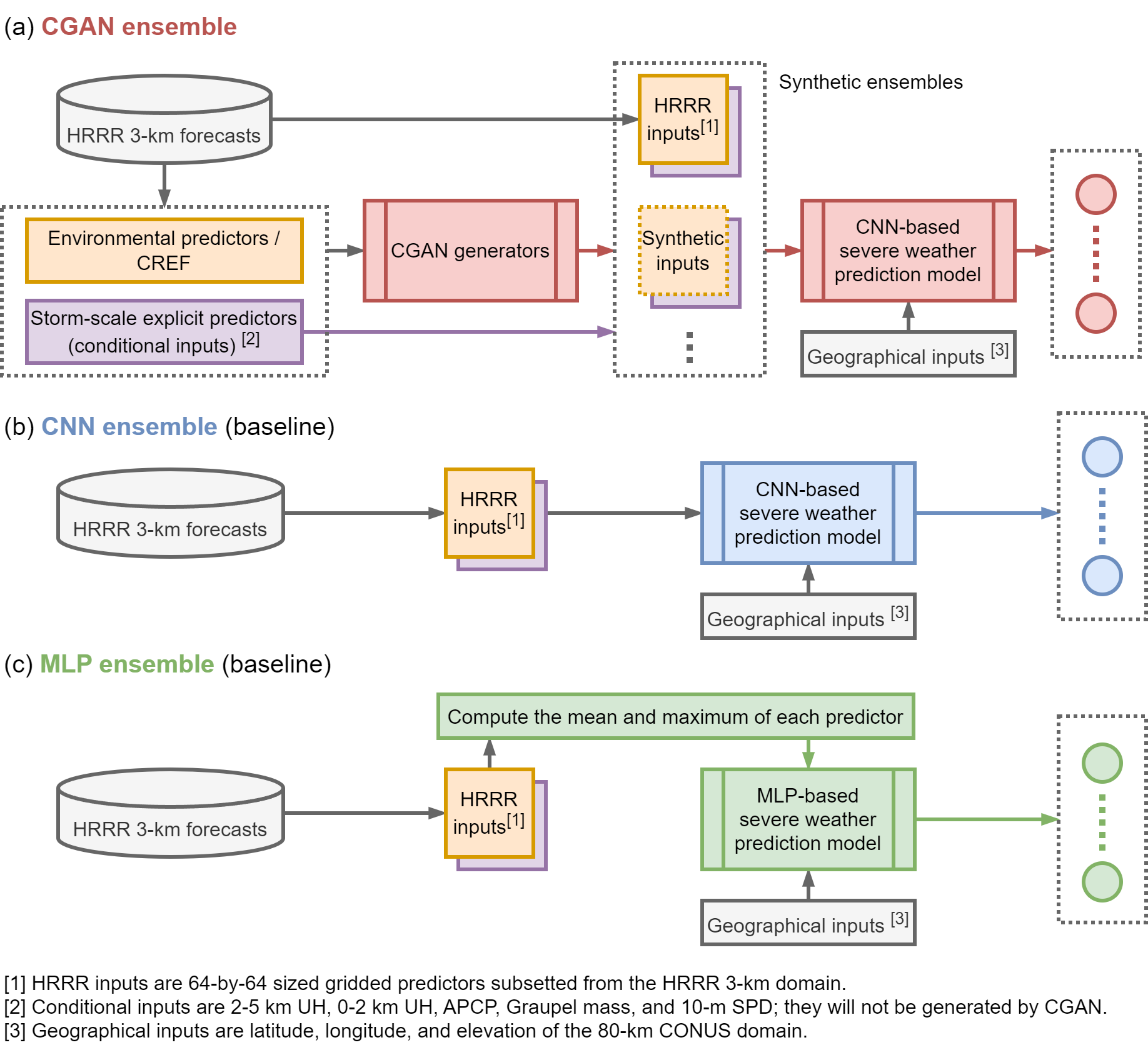}\\
 \caption{Technical steps of the CGAN ensemble (a), CNN ensemble (b), and MLP ensemble (c).}\label{fig2}
\end{figure}

Two neural-network-based post-processing steps were combined as the main methodology of this research. First, Conditional Generative Adversarial Networks (CGANs), a type of deep generative model, were applied to create synthetic CREF and environmental predictors. The CGAN-generated fields were paired with the original 2-5 km UH, 0-2 km UH, APCP, GRPL, and 10-m SPD fields, converting deterministic HRRR predictors into a set of pseudo-ensembles. Second, a CNN-based prediction system was applied; it provides severe weather probability estimations from each ensemble member independently. Combining these two steps would lead to ensemble predictions of severe weather. Note that the methodology of this research is not restricted to deterministic CAM forecasts only. In situations where an ensemble of CAM runs is available, the methods can be applied to expand the ensemble size.
 
The above post-processing method was trained from 5 July 2018 to 31 December 2020 with HRRRv3 and HRRRv4 forecasts as inputs and SPC reports as targets. It produces 4-hr, 80-km severe weather probabilities out to 24 hours with the exceptions of 00Z and 01Z; these two forecast lead times were ignored because of the HRRR model spin-up. The validation set was a 10\% random split from the training set, and it was fixed for all the training steps. The verification period of the final post-processing outputs is 1 January 2021--31 December 2021 and with HRRRv4 only. 

\subsection{Conditional Generative Adversarial Networks}\label{sec31}

\begin{figure}[t]
\noindent\includegraphics[width=\textwidth]{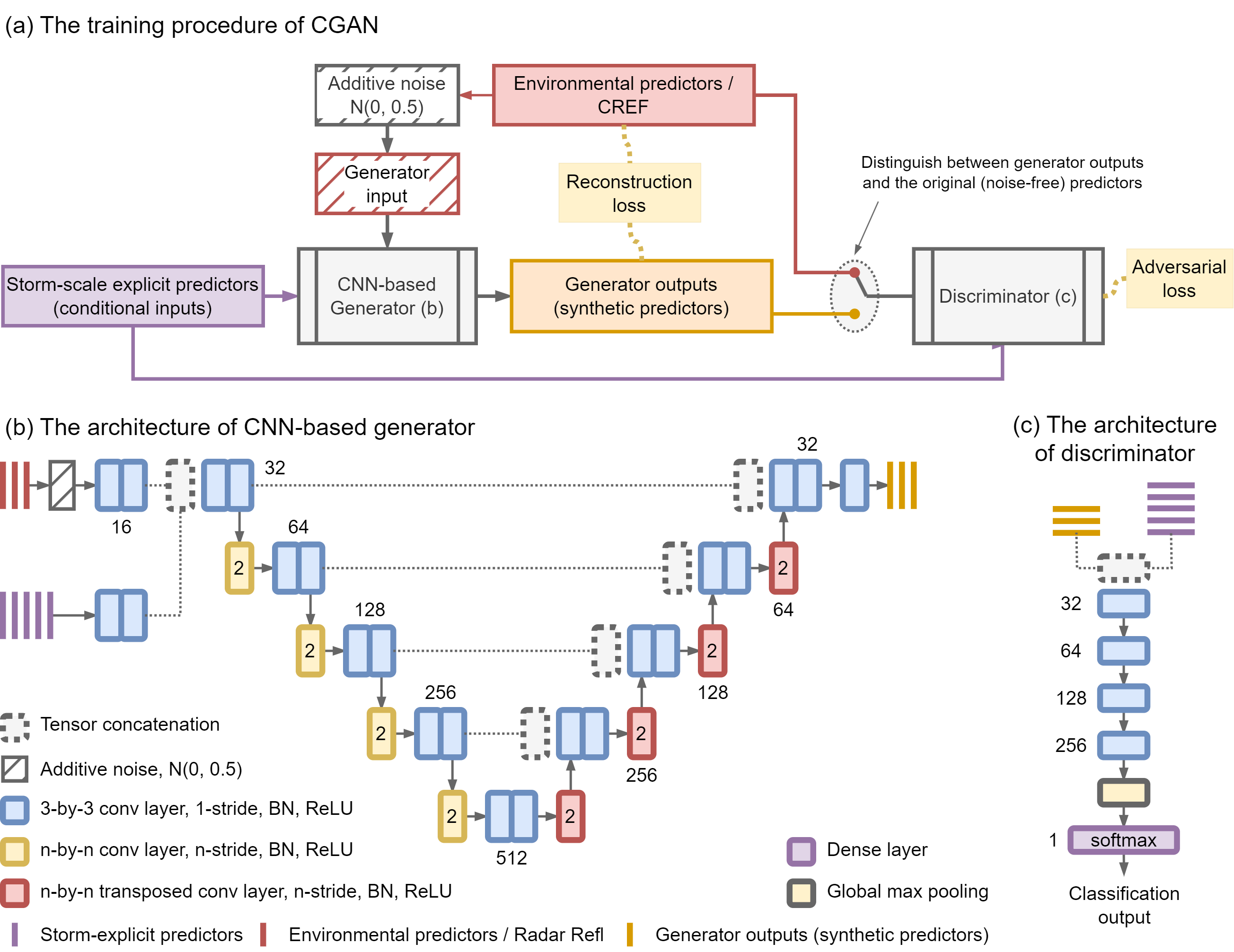}
\caption{(a) The training procedure of CGAN. (b) The architecture of the CNN-based generator; it contains convolutional layers (``conv''), Rectified Linear Unit (ReLU) activation function, and Batch Normalization (BN). Numbers beside each block represent the number of convolution kernels. (c) similar to (b), but for the architecture of the discriminator. Note that the CNN-based generator (b) will be preserved after training, whereas the discriminator is used for the CGAN training only.}\label{fig3}
\end{figure}

This research applies CGANs to generate synthetic ensembles from deterministic HRRR forecasts. CGANs are Generative Adversarial Networks (GANs) that utilize conditional information as additional inputs to enable more control over the sample generation process \citep{mirza2014,isola2017}. Here the conditional information is provided by five storm-scale explicit predictors: 2-5 km UH, 0-2 km UH, APCP, GRPL, and 10-m SPD. By accepting such conditional inputs, CGANs were trained to generate synthetic CREF and environmental predictors. The idea behind this choice is that UH and CREF are the two major CAM outputs that are related to convective storms directly \citep{clark2012,gallo2021}. If both UH and CREF are allowed to be generated by CGANs freely, the resulting synthetic ensembles may contain unrealistic storms and bias the severe weather prediction step. Thus, training CGANs to create CREF, but letting the generation process be conditioned on UH and other explicit predictors would balance the creativity and realism of CGAN outputs.

The CGAN in this research consists of a CNN-based generator and discriminator (Fig.~\ref{fig3}a). The generator has two input branches, one takes conditional inputs, and the other accepts the initial state of predictors to be generated. Such initial states were prepared by combining the original HRRR predictors with $\mathcal{N}(0, 0.5)$ distributed Gaussian noise (Fig.~\ref{fig3}b). Negative values will be corrected to zeros for the initial states of CAPE, CIN, and CREF. The base architecture of the generator is an encoder-decoder CNN with skip connections on each encoding level, known as UNET \citep{ronneberger2015}. The UNET applies convolutional layers with two strides and transposed convolutional layers for down- and up-samplings, respectively. The number of convolution kernels doubles after each down-sampling operation, from 32 to 512, and the numbers are symmetrical on the upsampling side (Fig.~\ref{fig3}b). The discriminator of the CGAN is a CNN-based binary classifier that aims to distinguish generator outputs from their corresponding HRRR predictors. Besides the classification inputs, the discriminator also takes the same conditional inputs as that of the generator (Fig.~\ref{fig3}c).

The optimization objective of CGAN is expressed as follows:

\begin{equation}\label{equ1}
\begin{array}{l}
\displaystyle\min_{G}{\max_{D}{\mathrm{\, CGAN\left(G,D\right)}}}=\mathcal{L}_{A}\left(G,D\right) + \lambda \mathcal{L}_{R}\left(G\right)\\
\mathcal{L}_{A} = \mathbb{E}_{x\sim p_x}\log{D\left(x|m\right)}+\mathbb{E}_{z\sim p_z}\log{\left[1-D\left[G\left(z|m\right)\right]\right]} \\
\mathcal{L}_{R} = \mathbb{E}_{x\sim p_x,\, z\sim p_z}\left\|x-G\left(z|m\right)\right\|_{1}
\end{array}
\end{equation}

\noindent
Where $G$ and $D$ are the CGAN generator and discriminator, respectively. $\mathcal{L}_{A}$ is the adversarial loss, and $\mathcal{L}_{R}$ is the reconstruction loss. The relative importance of the two loss functions is adjusted by $\lambda$. This research selects $\lambda=1$. $m$ is the conditional input. $x$ and $p_x$ represent the HRRR predictors to be generated, and their probabilistic distributions, respectively. $z$ is the initial state of $x$. $p_z$ is the probabilistic distribution of $z$. The generator and discriminator accept $x$, $z$, and $m$ as 64-by-64-sized input frames. $\left\|\cdots\right\|_{1}$ is L1 vector norm.

Fig.~\ref{fig3}a explains the training procedure of CGAN. In each iteration, the discriminator is updated first; its weights are optimized to maximize $\mathcal{L}_{A}$. Then the generator is updated by minimizing the combination of $\mathcal{L}_{A}$ and $\mathcal{L}_{R}$. The optimization of $\mathcal{L}_{A}$ introduces a competition between the discriminator and generator. With the generator trying to minimize the objective that the discriminator aims to maximize, it would generate realistic fields that look similar to HRRR predictors and cannot be distinguished easily. $\mathcal{L}_{R}$ is an additional loss function that regularizes the generator. This loss function compares the grid-point-level difference between generator outputs and their corresponding HRRR predictors; it is implemented to prevent the generator from amplifying the input noise.

Two CGANs of the above were proposed: one generates CAPE, CIN, and CREF (i.e., three input channels), and the other generates MSLP, 2-m Temp, 2-m Dewpoint T, 0-1 km SRH, 0-3 km SRH, 0-6 km U Shear, 0-6 km V Shear (i.e., seven input channels). The same CGANs were applied to all forecast lead times.

\subsection{CNN-based end-to-end severe weather prediction}\label{sec32}

\begin{figure}[t]
 \noindent\includegraphics[width=\textwidth]{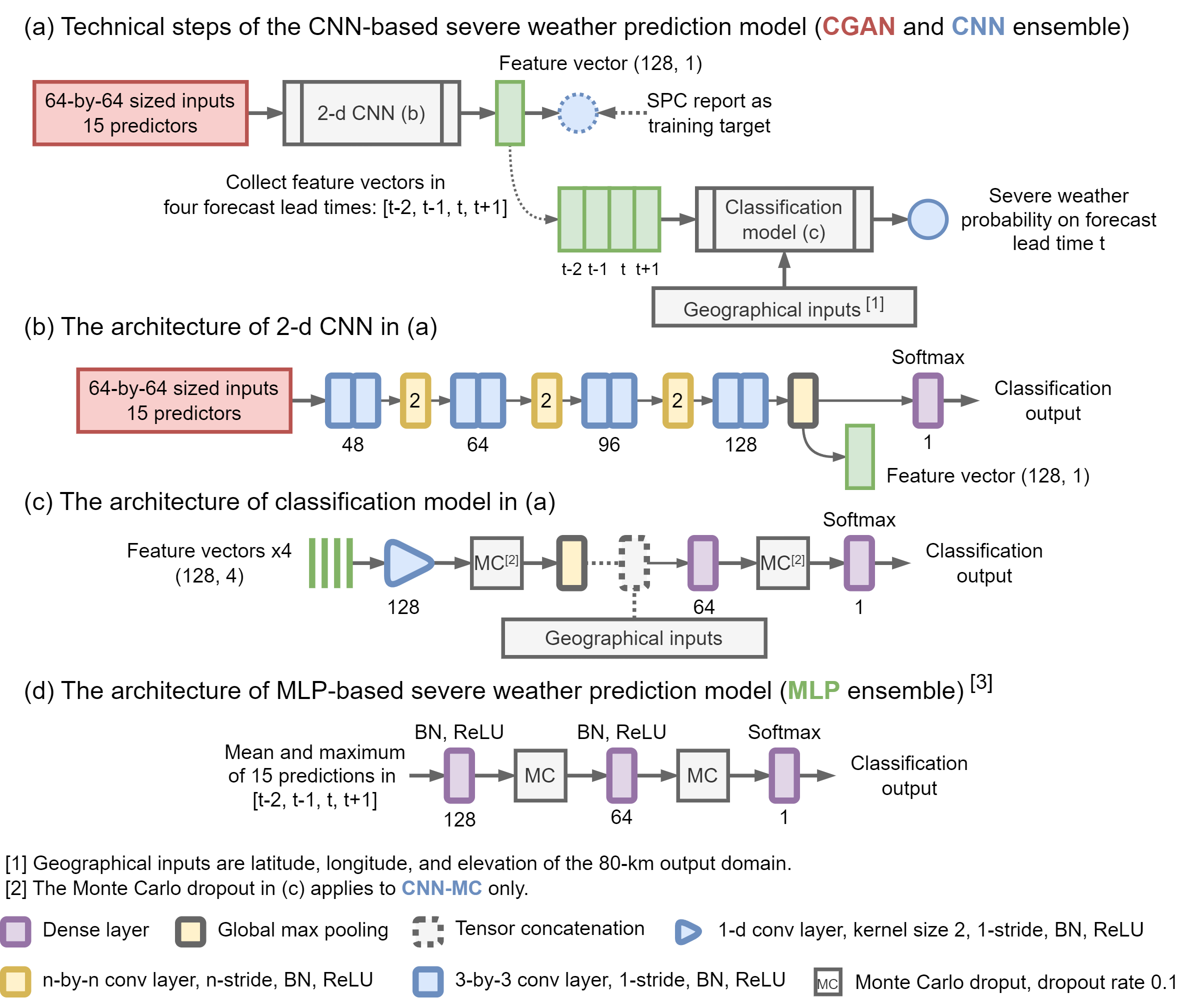}\\
 \caption{(a) The CNN-based end-to-end severe weather prediction model. (b) the architecture of the 2-d CNN in (a); it contains convolutional layers (``conv''), Rectified Linear Unit (ReLU) activations, and Batch Normalization (BN). (c) The architecture of the classification model in (a). (d) The architecture of the MLP-based severe weather prediction model. Monte Carlo (MC) dropout layers are included in (c) and (d)}\label{fig4}
\end{figure}

A CNN-based prediction model was applied to produce severe weather probabilities from 64-by-64-sized inputs from either HRRR forecasts or generated by the CGANs. The prediction model has two components: a representation learning model and a classification model (Fig.~\ref{fig4}a). The representation learning model is a CNN with 2-dimensional (2-d) convolution kernels; it compresses gridded inputs on individual forecast lead times and produces feature vectors that can represent severe-weather-related information (Fig.~\ref{fig4}b). The base architecture of the 2-d CNN is organized with two same-padding convolutional layers followed by a convolutional layer with two strides. The number of convolution kernels increases after every strided convolutional layer, from 48 to 128. The last layer of the 2-d CNN performs 2-d global max-pooling, which converts (8, 8, 128) sized tensors to a feature vector of (1, 128). The same 2-d CNN is applied to inputs of all forecast lead times.

The classification model takes four feature vectors produced by the representation learning model inputs (for 00Z and 01Z forecasts, it takes 2 and 3 feature vectors, respectively, because of the HRRR model spin-up). The classification model begins with a 1-d convolutional layer to process the forecast lead time dimension. The resulting tensor is passed through a 1-d global max-pooling layer and concatenated with normalized geographical inputs of latitude, longitude, and elevation. Two dense layers are applied after the concatenation step and produce severe weather probabilities using a sigmoid function (Fig.~\ref{fig4}c). Classification models were trained individually on each forecast lead time window. For example, classification model weights for 02-06Z and 03-07Z forecasts were different.

Monte Carlo (MC) dropout \citep{gal2016} was implemented within the classification model for ensemble prediction and uncertainty quantification (Fig.~\ref{fig4}c, d). MC dropout deactivates part of the neurons randomly, similar to the conventionally used dropout method \citep{srivastava2014}, but works for both training and inference stages. Thus, by running the same classification model multiple times, different subsets of neurons would be deactivated, and the remaining neurons can be viewed as a slightly different neural network. This adds stochasticity to the classification models and enables them to make ensemble predictions. Note that the MC dropout was applied to the classification model only; the representation learning model ((Fig.~\ref{fig4}a) was not affected.

The technical highlight of the CNN-based severe weather prediction model is the decoupling of representation learning and classification. Existing studies have shown that such decoupling can improve the performance of neural networks on long-tailed classification problems \citep[e.g.][]{kang2019}, which is indeed the case for severe weather predictions. Further, the separate representation learning model does not need to handle inputs from multiple forecast lead times at once, and thus, its complexity and computational costs can be reduced (e.g., 3-d convolution kernels are not needed).

The representation learning model was trained from 5 July 2018 to 31 December 2020, using HRRRv3 and v4 forecasts. SPC reports on the corresponding forecast lead times were used as training targets. This is achieved by configuring an additional dense layer with a sigmoid activation function to compute cross-entropy loss (Fig.~\ref{fig4}b). 

\subsection{Hyperparameter optimization and baseline methods}\label{sec33}

The main method of this research combines the CGAN-generated ensembles and the CNN-based end-to-end severe weather prediction model. Hereafter, this system is referred to as the ``CGAN ensemble'' (Fig.~\ref{fig2}a). Two baseline methods, ``CNN ensemble'' and ``MLP ensemble'' were applied to compare against the CGAN ensemble. The CNN ensemble features the same CNN-based severe weather prediction model as the CGAN ensemble, but without using CGANs. That said, this baseline produces ensembles from MC dropouts only. By comparing the performance between the CGAN ensemble and the CNN ensemble, the contribution of CGANs can be measured.

The MLP ensemble does not rely on CNNs to make severe weather predictions. Instead, it combines an MLP with feature engineering steps. Given the same 64-by-64-sized HRRR predictors that CNN-based methods would use, the MLP baseline converts them into scalar inputs by computing their mean and maximum values on spatial and forecast lead time dimensions. This pre-processing step is similar to \citet{sobash2020}.  The architecture of MLP is illustrated in Fig.~\ref{fig4}d; it consists of two hidden layers with BN, Rectified Linear Unit (ReLU) activation function, and MC dropout. Several existing works have shown that MLPs are effective in deriving severe weather probabilities from CAM outputs \citep[e.g.][]{sobash2020}. Notably, they can outperform the conventionally used surrogate methods. Thus, in this experiment, the MLP baseline is a fair representation of other machine-learning-based severe weather models that would be used in an ensemble prediction setup. By comparing its performance with the two CNN-based methods, the contribution of CNN-based severe weather predictions can be examined.

Hyperparameter optimizations were conducted on the main method and the two baselines. For each neural network, the relative importance of its hyperparameter options was identified based on the prior knowledge of the network design and other existing studies \citep[e.g.][]{greff2016lstm,sharma2019hyperparameter,yu2020hyper}. Experiment trials were then proposed from the most important to the least important options. The performance criteria of the CGAN trials were the reconstruction loss and stability. Low reconstruction loss across all the generated predictors was preferred. For the CNN-based representation learning model, CNN-based classifier, and the MLP baseline, their performance criteria were severe weather prediction skills averaged across all forecast lead times. The hyperparameter optimization of this study was conducted based on the same training and validation set, and without accessing the verification data. For the two CGANs, each trial was fixed to 200 epochs. For the CNN-based representation learning model, CNN-based classifier, and the MLP baseline, their trials had early stopping implemented. At the end of the hyperparameter search, neural network weights from the best trials were preserved.

The results of hyperparameter optimization for all models, from the most important to the least important options are provided as follows:

\begin{itemize}
    \item CGAN: $\mathcal{N}(0, 0.5)$ additive noise, 1e-4 learning rate with decay, batch size 32, Adam optimizer \citep{kingma2014}, ReLU activation (see Fig.~\ref{fig3}b).
    
    \item CNN-based representation learning model: sample rebalancing of 1 positive: 10 negatives, 1e-4 learning rate with decay, batch size 32, Adam optimizer, ReLU activation, \{48, 64, 96, 128\} number of convolution kernels and 2 convolutional layer stacks per downsampling block, global max pooling (see Fig.~\ref{fig4}b).

    \item CNN-based classifier: sample rebalancings of 1:1, dropout rate 0.1, 1e-4 learning rate with decay, batch size 32, Adam optimizer, ReLU activation, 128 1-d convolution kernels and 64 dense layer neurons (see Fig.~\ref{fig4}c). 

    \item MLP baseline: sample rebalancings of 1:1, dropout rate 0.1, 1e-4 learning rate with decay, batch size 64, Adam optimizer, ReLU activation, 128, and 64 neurons for the two dense layers (see Fig.~\ref{fig4}d).
    
\end{itemize}

The model training and hyperparameter optimizations of this study were conducted on NVIDIA Tesla V100 GPUs with additional CPUs to support the data pipeline. The MLP baselines were roughly 30\% faster than the CNN baseline and 70\% faster than the CGAN method.

\subsection{Verification methods}\label{sec34}

To assess forecast quality from the three different ML-based prediction systems, we verify the ensemble mean severe weather probability against SPC reports from 1 January 2021--31 December 2021 using Brier Score (BS) and Brier Skill Score (BSS; \citet{murphy1973}). The climatology reference of BSS was derived from SPC reports between 1986 and 2015, and separately for locations, day of the year, and hour of the day. The spatial or temporally aggregated BSSs were computed as follows: given gridded severe weather probability values on the CONUS 80-km domain, the BSs were computed first for individual initialization days, forecast lead times, and grid cells. Then, the resulting three-dimensional arrays were averaged temporally or spatially. Finally, the climatology reference was averaged in the same way to produce BSSs. The above steps follow the suggestion of \citet{hamill2006}. Three-component decomposition of BSs and reliability diagrams were also computed to attribute the BSS difference; their computation follows \citet{murphy1973} and \citet{hsu1986}.

The uncertainty quantification of the three methods was evaluated using spread-skill diagrams and discard tests. The spread-skill diagram compares the standard deviation of ensemble members against their root mean squared prediction errors. Spread-skill reliability can be computed from this diagram as a measure of the correlation between the uncertainty and the spread of a given ensemble \citep{delle2013,haynes2023}. The discard test is a step-wise evaluation that shows how model errors would change when the current most uncertain predictions are removed. Monotonicity fraction can be derived from the discard test results; it measures the correlation between the uncertainty and the predictive error of a given ensemble \citep{barnes2021,haynes2023}.

The performance of the CGAN ensemble in Section \ref{sec3}\ref{sec31} will be examined in two ways. First, pattern correlations \citep{murphy1989} between CAPE and CIN, CREF and 2-m Dewpoint T, and 0-1 km and 0-3 km SRH were computed over the HRRR domain on each initialization day and forecast lead time, and separately for CGAN-generated predictors and their HRRR counterparts. This comparison demonstrates how well CGANs preserve inter-variable correlations. Second, the permutation feature importance \citep{altmann2010} was estimated from the pre-trained CNN severe weather model in Section \ref{sec3}\ref{sec32} using either synthetic predictors or HRRR predictors. The resulting feature importance differences indicate how CGAN outputs would impact the decision-making of the CNN-based prediction model.

\section{Results}\label{sec4}
\subsection{Case-based assessments}\label{sec41}

\begin{figure}[t]
 \noindent\includegraphics[width=\textwidth]{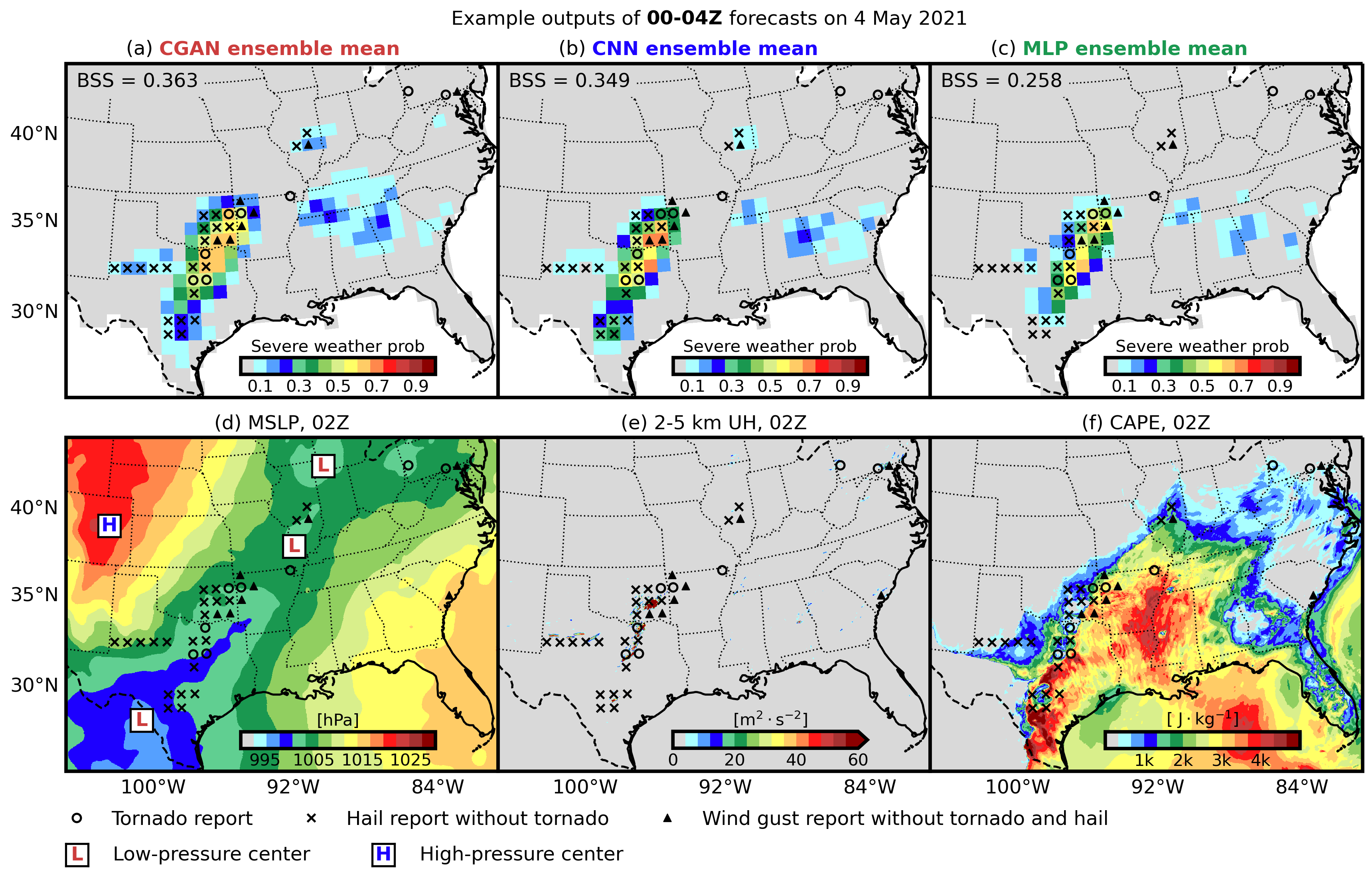}\\
 \caption{Severe weather predictions for 0000-0400 UTC 4 May 2021. (a-c) Severe weather probabilities produced by the CGAN, CNN, and MLP ensemble mean, respectively. (d-f) HRRRv4 forecasts of MSLP, 2-5 km UH, and CAPE. The forecasts were initialized on 0000 UTC 4 May 2021.}\label{fig5}
\end{figure}

Case-based assessments are presented to demonstrate severe weather predictions of the three different post-processing methods. In Fig.~\ref{fig5}, an example is provided where all methods performed well.

The HRRRv4 is initialized on 0000 UTC 4 May 2021, with its 2-hr forecast fields of MSLP, 2-5 km UH, and CAPE shown in Fig.~\ref{fig5}d, e, and f, respectively. Based on the MSLP field, a frontal system is present across the Central Plains, with environmental conditions favorable for convective severe weather (Fig.~\ref{fig5}d). From 0000 to 0400 UTC, 35 severe reports were received, with 6 tornadoes, 8 reports of wind gusts, and 21 reports of either hail or hail combined with wind gusts. Most of these cases were located along the boundary of cold and warm air masses, which can be identified by the transition from near-zero (stable atmosphere controlled by the cold air mass) to high-positive (convection-favored atmosphere controlled by the warm air mass) CAPE values (Fig.~\ref{fig5}f). A few grid cells with high-positive 2-5 km UH are found within the HRRR forecast, suggesting the existence of supercells. Several hail reports and 2 tornadoes occurred within the high UH area.

Post-processing methods take HRRRv4 forecasts on 0200 and 0300 UTC 4 May 2021 as inputs and produce 4-hr, 80-km severe weather probabilities for 0000-0400 UTC (0000 and 0100 UTC forecasts were ignored because of spin up. See Section \ref{sec3}). For CONUS grid cells that contain high UH HRRR forecasts, all three methods agreed well, with probabilities up to 60\% (Fig.~\ref{fig5}a--c).

Quantitatively, the BSSs of the two CNN-based methods were higher than that of the MLP baseline, with the CGAN ensemble having the highest BSS. This BSS difference is primarily explained by the ability of each method to identify severe weather cases that were not co-located with large magnitudes of 2-5 km AGL UH. 12 CONUS grid cells contained severe weather reports, but the corresponding HRRR 2-5 km UH was lower than $\mathrm{10 m^2s^{-2}}$. The MLP baseline performed poorly at these grid cells, whereas the two CNN-based methods did better (c.f. severe weather reports in Arkansas and Illinois in Fig.~\ref{fig5}a--c). Moreover, the CGAN ensemble produced more grid cells with severe weather probabilities $>=$ 0.1 (Fig.~\ref{fig5}a; cyan-colored shades). These grid cells are located around the intersection of cold and warm air masses, which are locations that favor mesoscale convection and severe weather. That said, when explicit predictors, such as 2-5 km UH, are forecasted correctly and contain clear signals, all methods can predict severe weather probabilities well. On the other hand, when severe weather cases do not co-locate with high UH values or forecasted supercells, the CNN-based methods, especially the CGAN ensemble, showed a stronger ability to predict severe weather by identifying specific patterns from environmental predictors (e.g., CAPE), which leads to better performance overall.

\begin{figure}[t]
 \noindent\includegraphics[width=\textwidth]{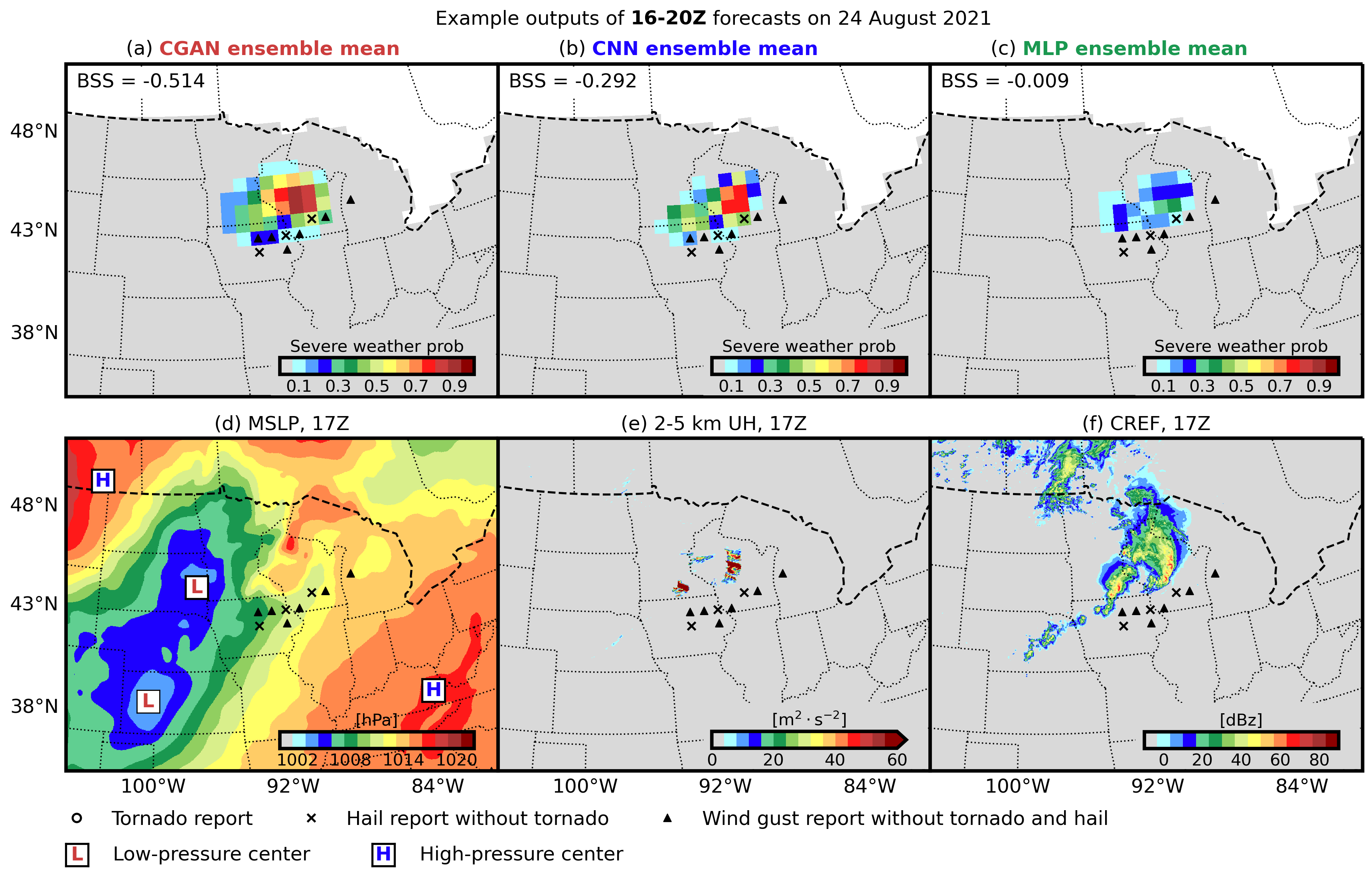}\\
 \caption{As in Fig.~\ref{fig5}, but for forecasts valid 1600--2000 UTC 24 August 2021.}\label{fig6}
\end{figure}

In Fig.~\ref{fig6}, another case assessment is presented where all the methods, especially the CGAN ensemble, performed poorly. The case is a frontal cyclone system that originated from the Pacific Ocean. On 21 August, the system appeared in the Pacific Northwest and started moving eastward. On 24 August, when the case assessment began, the warm front of the system approached the Midwest; it triggered a set of thunderstorms, causing hail and wind gust damage in the local area.

The overall development of this synoptic system was captured by the HRRRv4 forecasts. However, in the 17-hr forecasts in Fig.~\ref{fig6}d--f, there is a positional error. The surface low is displaced to the east and warm air mass to the north in the forecast. As a result, the forecasted 2-5 km UH and CREF patterns do not match the location of SPC reports. All the methods were biased by this positional error, and the highest probabilities were to the north of the SPC reports (Fig.~\ref{fig6}a-c). This leads to the double penalty of misses and false alarms. The CGAN ensemble performed the worst; it produced the highest severe weather probabilities where no severe weather cases were reported, causing -0.514 BSS.

This example demonstrates that the underlying skill of HRRR forecasts has a strong influence on the ML-based severe weather post-processed forecasts. CNN-based methods are more sensitive to certain input patterns related to severe weather. This ability helps detect potential severe weather cases when the HRRR forecasts inputs are skillful. When the HRRR forecasts have large positional errors, CNN-based methods will likely be penalized for overconfidence. In addition, the CGANs can amplify such positional errors because their sample generation process is conditioned on explicit predictors of the original HRRR forecasts. This explains why the CGAN ensemble issued the highest severe weather probabilities on the biased locations. We have examined all cases from 1 January 2021--31 December 2021 and can confirm that, despite a few worst-case scenarios like Fig.~\ref{fig6}, most of the HRRRv4 forecasts were skillful enough to support severe weather post-processing.

\subsection{Brier skill scores and reliability diagrams}\label{sec42}

\begin{figure}[t]
 \noindent\includegraphics[width=\textwidth]{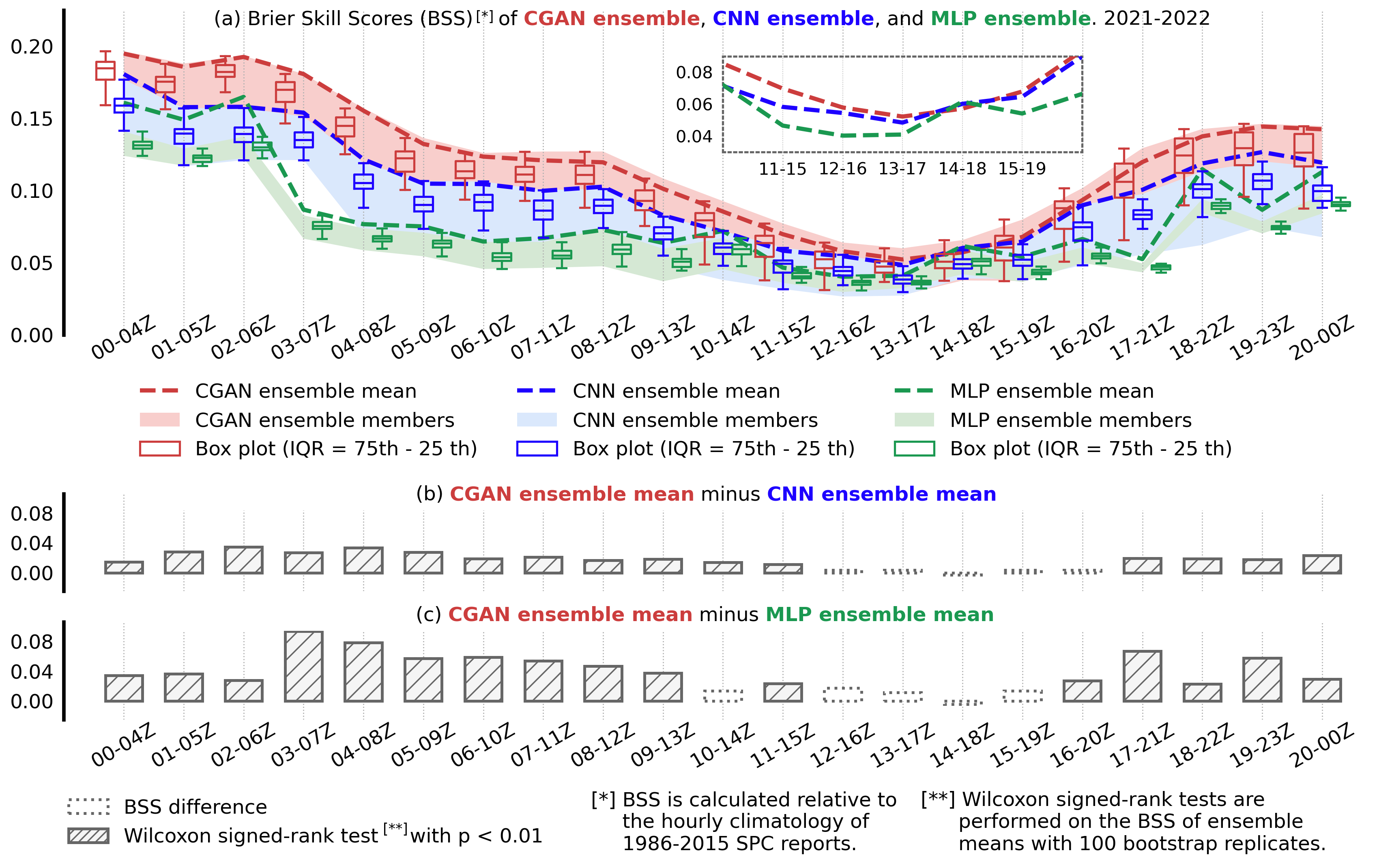}\\
 \caption{Verification of post-processed hourly severe weather probabilities with Brier Skill Scores (BSSs; higher is better) by forecast lead time. (a) BSS curves for the CONUS domain, initializations from 1 January 2021--31 December 2021, and the CGAN, CNN, and MLP ensembles. Solid dashed lines represent the BSS of the ensemble mean. Shaded areas and boxplots represent the BSSs of individual ensemble members. The Inter Quantile Range (IQR) of boxplots are 25th and 75th percentiles. Inset plot shows the BSSs from 11-15Z to 15-19Z. (b) BSS difference between CGAN and CNN-MC ensemble mean. (c) BSS difference between CGAN and MLP ensemble mean.}\label{fig7}
\end{figure}

The BSSs were aggregated over the CONUS for each 4-hr forecast lead time window (Fig.~\ref{fig7}). All three methods exhibited similar BSS variations (Fig.~\ref{fig7}a). BSSs were largest between 00-06Z, with the CGAN ensemble reaching 0.2 and the two baselines staying around 0.16. From 12-18Z, the BSSs of all methods decreased to their day-1 minimums, ranging from 0.04 to 0.06. Beyond 18Z, the BSSs of all three methods increased, with the CGAN ensemble mean approaching 0.15 and the two baselines near 0.12. In addition, the ensemble mean of all methods showed higher BSSs compared to most of the individual members, indicating that the ensemble ML predictions of severe weather was beneficial for improving forecast skills from deterministic predictions (Fig.~\ref{fig7}a).

All three methods generated forecasts with relatively low skill during the overnight through early morning hours (i.e., 09--15Z; (Fig.~\ref{fig7}a)). Several factors could be contributing to the relatively poor forecast skill. Fewer severe weather events are typically reported overnight, and thus, less training information is available. Further, overnight severe weather could be related to nocturnal convection that is often more challenging to forecast \citep[e.g.][]{johnson2017,blake2017}. Errors in the input HRRR predictions may lead to poor post-processed forecasts, especially if errors are not biased in a systematic way that can be corrected by the ML algorithms.

For other forecast lead times, the CGAN ensembles outperformed the two baselines with statistically significant BSSs increases (Fig.~\ref{fig7}b,c) . The performance gains from the CNN baseline to the CGAN ensemble indicate that the CGANs contributed positively to the CNN-based prediction model. The CNN baseline and MLP baseline showed comparable BSS performance, with the CNN baseline being slightly better, especially for 06-12Z (Fig.~\ref{fig7}a).

\begin{figure}[t]
 \noindent\includegraphics[width=\textwidth]{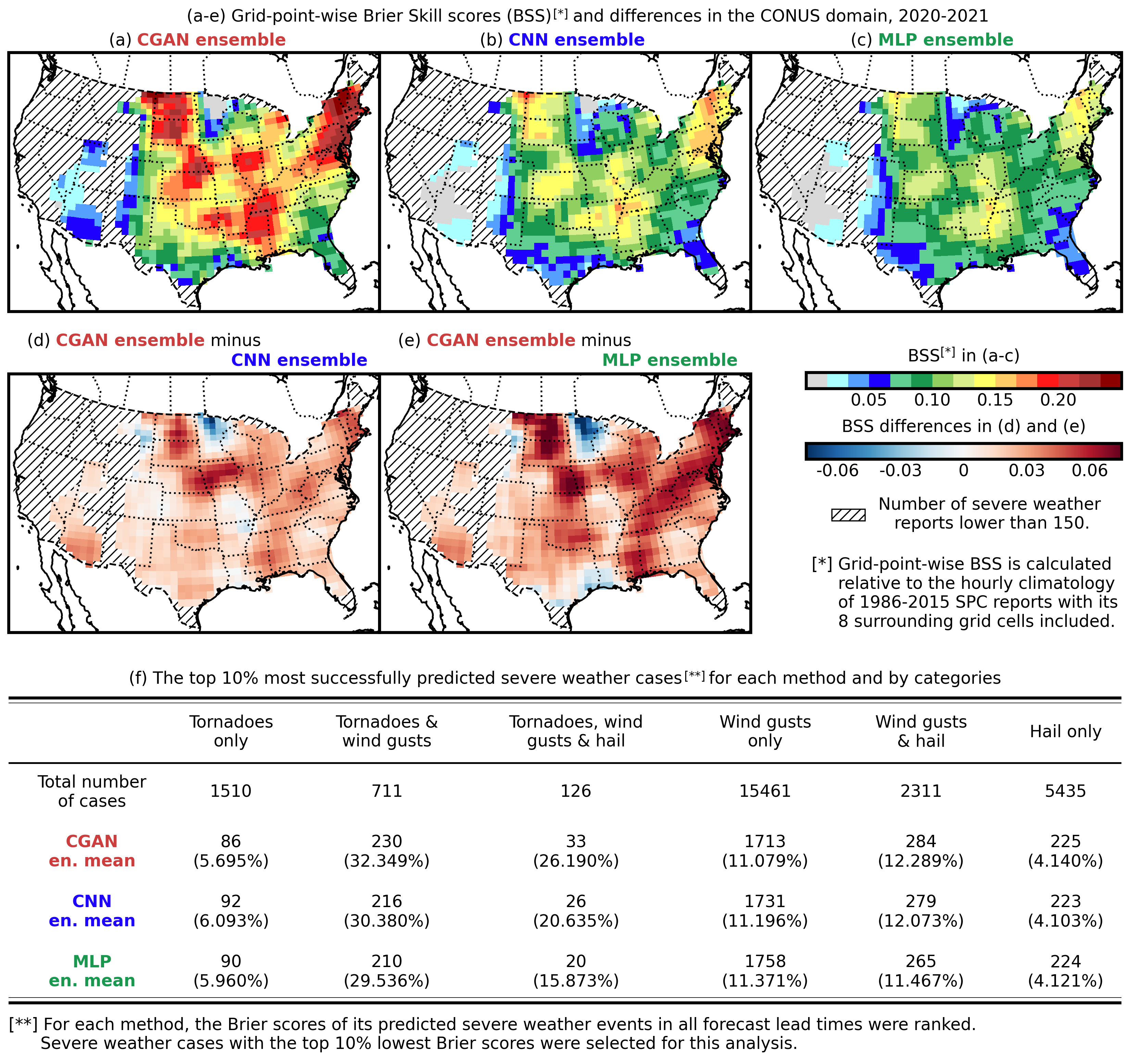}\\
 \caption{(a) Grid-point-wise BSSs computed using the CGAN ensemble mean for all initialization days and forecast lead times. (b) As in (a), but for the CNN ensemble mean. (c) As in (a) but for the MLP ensemble mean. (d) is the difference between (a) and (b). (e) is the difference between (a) and (c). Neighboring grid cells are included in the computation of grid-point-wise BSSs. Grid cells with severe weather reports lower than 150 are marked using hatches. (f) The categorical distributions of the top 11.45\% most successfully predicted severe weather cases from the CGAN ensemble mean, the CNN ensemble mean, and the MLP ensemble mean.}\label{fig8}
\end{figure}

Grid-point-wise BSSs were computed using all initialization days and forecast lead times. For each CONUS grid cell, verification results from its surrounded 3-by-3 grid cells were included to increase the sample size. The West Coast, Pacific Northwest, and some boundary grid cells do not have enough SPC reports, and their BSSs are not shown.

All methods performed well in the northeastern United States and the Northern and Central Plains. On the other hand, some areas in the Southeast, Florida, and southern Arizona exhibited lower forecast skill (Fig.~\ref{fig8}a-c). The good performance over the Great Plains is primarily explained by the connection between severe weather events and supercell thunderstorms, which are often well captured and forecast by explicit predictors such as 2-5 km UH.

In the northeastern U.S., all methods, especially the two CNN-based methods, performed well (Fig.~\ref{fig8}a,b). Most of the reports in this region are wind reports, which are often obtained from non-supercellular convective modes. The good performance of the two CNN-based methods shows that they can predict severe weather events without relying on UH. The good performance in the northeastern United States could be affected by regional bias, since many wind reports in this area do not occur in association with $>=$ 50 kt wind gusts \citep[e.g.][]{sobash2020}, which makes severe weather reports occur more often than they should. Post-processing methods can overfit this regional bias by using latitude and longitude as predictors.

Southern Arizona and the Southeast exhibit poor performance. A possible reason is that the severe weather reports in these two areas could be associated with monsoon or "pulse"-type thunderstorms. These storm modes are generally short-lived and more difficult to forecast than supercells, resulting in limited forecast skills. In addition, the coastal environment of the Southeast adds additional difficulties to the prediction of severe weather (e.g., see-breeze convection \citet{hock2022}).

For most parts of the CONUS domain, the CGAN ensemble outperformed the CNN baseline (Fig.~\ref{fig8}d), which in turn, outperformed the MLP baseline (c.f. Fig.~\ref{fig8}e). The CGAN ensemble and the CNN baseline rely on the same CNN for representation learning, which is trained using 64-by-64-sized inputs from the entire CONUS domain (see Section \ref{sec3}.\ref{sec32}). The good performance of the two CNN-based methods indicates that such representation learning is effective across various spatial and climatological conditions. In addition, the BSS increase from the CNN baseline to the CGAN ensemble further confirms that the CGANs have contributed positively to severe weather predictions.

Note that the CGAN ensemble performed worse than the two baselines in some grid cells. This was primarily due to the positional error of the HRRR forecasts. One of these cases is shown in Section \ref{sec4}.\ref{sec41}. We think that with CAM forecast skills being improved in the future, such double penalty problems will be substantially reduced.

The performance of severe weather prediction models varies by both regions and severe weather categories. In Fig.~\ref{fig8}f, the categorical distributions of correctly predicted severe weather events were provided for each method. The purpose of this analysis is to investigate which severe weather categories are more challenging to predict for each post-processing method. For severe weather events within the verification data, their individual Brier scores (i.e., the squared difference between the predicted probability and 1.0) associated with the ensemble mean of each post-processing method were ranked, with lower scores indicating more successful predictions. In this analysis, the top 10\% of the most successful predictions of each method were selected for each forecast lead time, and their severe weather categories were summarized.

From Fig.~\ref{fig8}, all methods tended to perform well in combined severe weather events. This is likely because these combined events have left stronger signals within the HRRR forecasts. For example, given 711 combined tornado and wind gust events, all methods captured at least 210 of them with good Brier scores. The combined tornadoes and wind gusts were likely caused by supercell thunderstorms; they created an ideal environment for tornado genesis and brought wind gust damage from downbursts. Supercell thunderstorms have clear connections with UH, and thus, the resulting severe weather events were predicted well by all methods. On the contrary, isolated events, especially ``tornadoes only'' and ``hail only,'' were less likely to achieve the top 10\% Brier scores. These isolated events could be relatively mild (i.e., non-supercell tornadoes) or occurred in regions where HRRR forecast skills were poor (e.g., hail reports on the West Coast and the Rockies), thus leading to somewhat unsatisfactory predictions. In summary, the predictability of each severe weather category has a larger impact than the choice of post-processing methods. Combined severe weather events (e.g., tornadoes and wind gusts) are more likely to be predicted successfully.

\begin{figure}[t]
 \noindent\includegraphics[width=\textwidth]{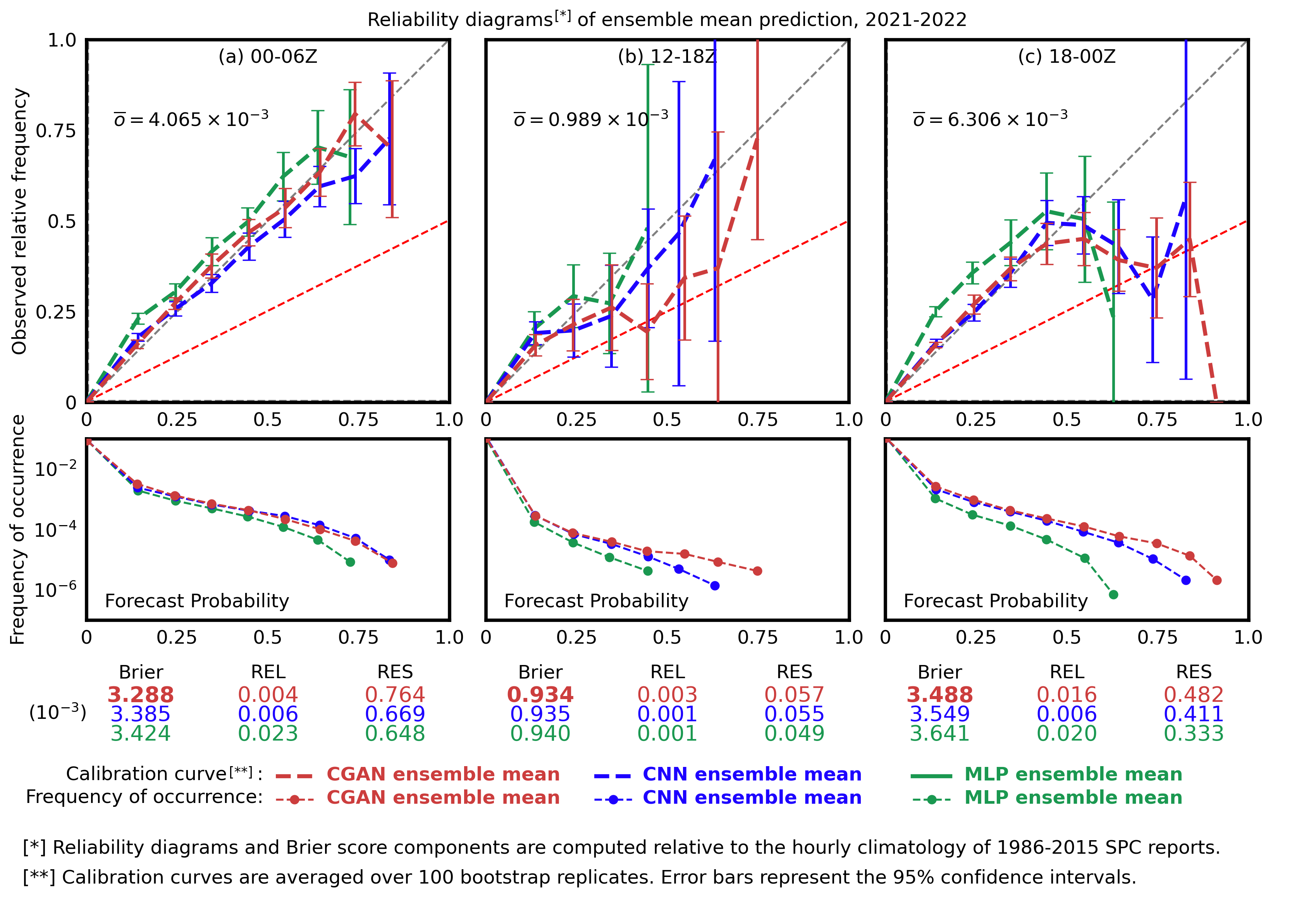}\\
 \caption{Verification of post-processed 4-hr severe weather probabilities with reliability diagrams, frequency of occurence plots, and Brier score (``Brier''; lower is better) decompositions [reliability (``REL''; lower is better), resolution (``RES''; higher is better), and climatological uncertainty ($\overline{o}$)]. All scores are displayed with a scale of $10^{-3}$. In (a-c), metrics are computed over 4-hr forecasts for 00-06Z, 12-18Z, and 18-00Z, respectively. No-skill reference lines and perfect reliability diagonal reference lines are included. Calibration curves are bootstrapped with 100 replicates, with their error bars representing the 95\% confidence intervals.}\label{fig9}
\end{figure}

Reliability diagrams in Fig.~\ref{fig9} provide further details regarding the ensemble mean performance of all methods in 00-06Z, 12-18Z, and 18-00Z. For low-probability forecasts, all methods had reliability curves aligned with the perfectly reliable reference lines, indicating that they predicted non-severe weather cases properly. This also explains the good reliability score overall, because more than 99\% of the verification cases were non-severe weather.

The two CNN-based methods produced more high-probability predictions. For 00-06Z, most of the predictions were accurate enough to separate conditional and unconditional observations, and thus, contributed to the resolution scores and BSs performance. For 12-18Z and 18-00Z, the high-probability predictions of the two CNN-based methods are somewhat overconfident. However, when compared to the MLP baseline, which generated fewer high-probability predictions, the two CNN-based methods were better in terms of resolution scores and BSs, because their reliability curves, although contain conditional bias, are still located within the positive-skill area (i.e., above the no-skill reference line). The CGAN ensemble showed clearly the best performance in the verification of 00-06Z and 18-00Z.  For these two forecast lead time groups, the CGAN ensemble produced as many high-probability predictions as the CNN baseline, but these predictions were more skillful than the latter, improving the resolution scores and BSs.

\subsection{Evaluations of uncertainty quantification}\label{sec43}

\begin{figure}[t]
 \noindent\includegraphics[width=\textwidth]{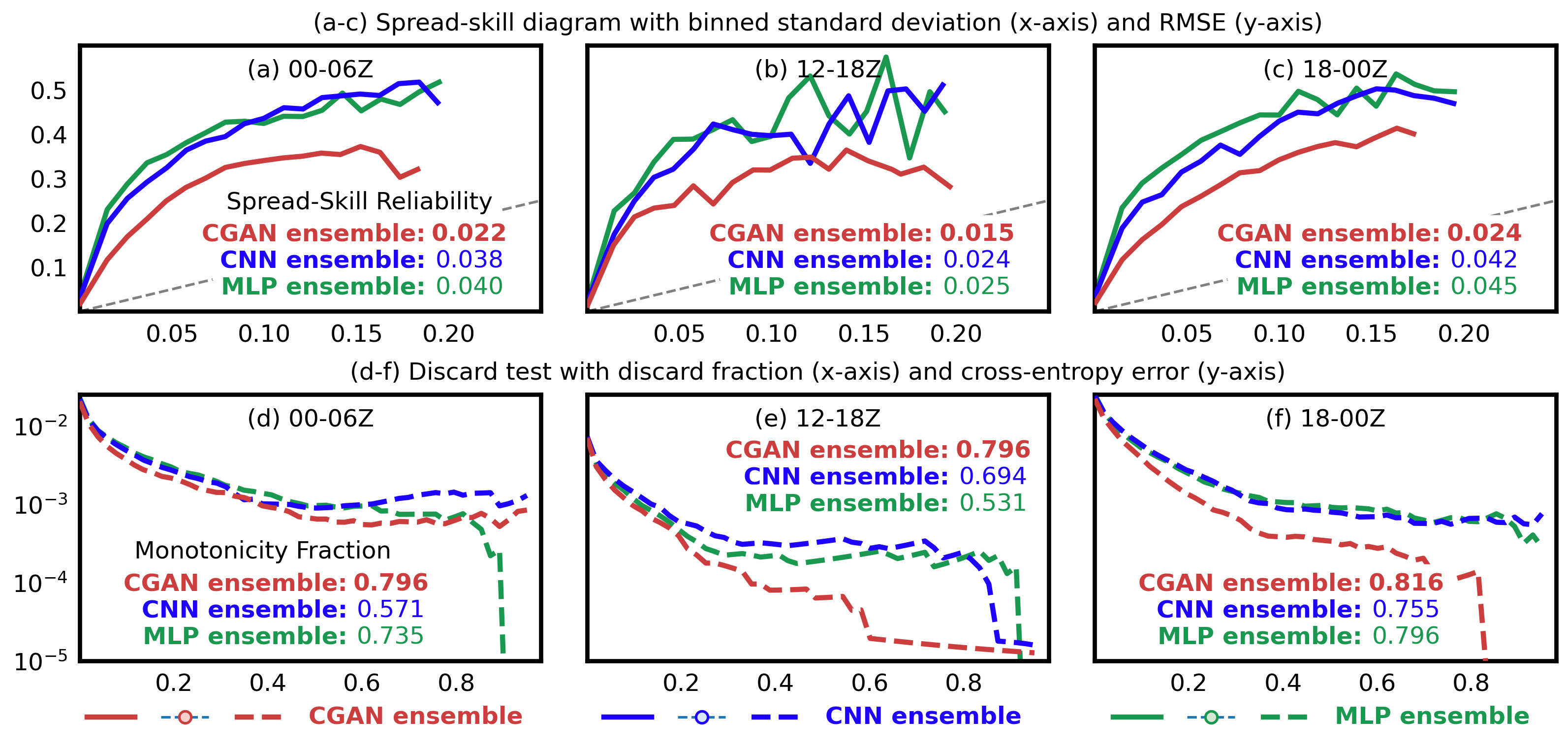}\\
 \caption{Verification of post-processed hourly severe weather probabilities with spread-skill diagram (a-c), and discard test curves (d-f). For (a-c), spread-skill reliability (lower is better) and perfect reliability diagonal reference lines are included. For (d-f), The Monotonicity Fraction (MF; closer to 1.0 is better) is provided. All metrics are computed over hourly forecasts for 00-06Z, 12-18Z, and 18-24Z.}\label{fig10}
\end{figure}

Spread-skill diagrams and discard tests were applied to evaluate the ensemble predictions of all methods in terms of their uncertainty quantification. Different from Section\ref{sec4}.\ref{sec42} which focuses on the forecast skill of the ensemble mean, here, the evaluations are conducted on individual ensemble members directly.

The quantitative relationships between the ensemble spread and the predictive errors are examined through spread-skill diagrams (Fig.~\ref{fig10}a-c). All methods have their spread-skill reliability lines staying above the perfectly reliable reference lines, indicating that their ensemble predictions are overconfident for uncertainty quantifications. That said, the ensemble spread is typically too narrow compared to the root mean squared error of the ensemble mean. Although not perfectly calibrated, the CGAN ensemble exhibited the best spread-skill reliability (Fig.~\ref{fig10}a-c). Compared to the two baselines which rely solely on MC dropouts to form ensembles, the participation of CGAN outputs covered more uncertainties that are potentially related to the variability of CREF and environmental predictors.

The discard test in Fig.~\ref{fig10}d-f examines the ranking quality of uncertainty estimates. In this evaluation, ensemble predictions are considered skillful if the overall prediction error of the ensemble mean, as measured by cross-entropy, decreases monotonically when fractions of the highest uncertainty verifications are discarded. The CGAN ensemble performed the best in terms of maintaining the monotonic decrease of its prediction error (MF closer to 1.0). Also, the prediction error of the CGAN ensemble decreased faster than the two baselines, pointing out that its uncertainty quantification better separated high- and low-uncertainty prediction regimes.

The combination of good discard test performance and poor spread-skill reliability performance suggests that all methods can rank the uncertainty of various prediction cases. However, they have underestimated the uncertainty of their predictions. That said, their ensemble spread would increase for more difficult prediction cases, but the amount of such increase is not scaled one-to-one for the prediction error.

\subsection{Evaluations of CGAN outputs}\label{sec44}

\begin{figure}[t]
 \noindent\includegraphics[width=\textwidth]{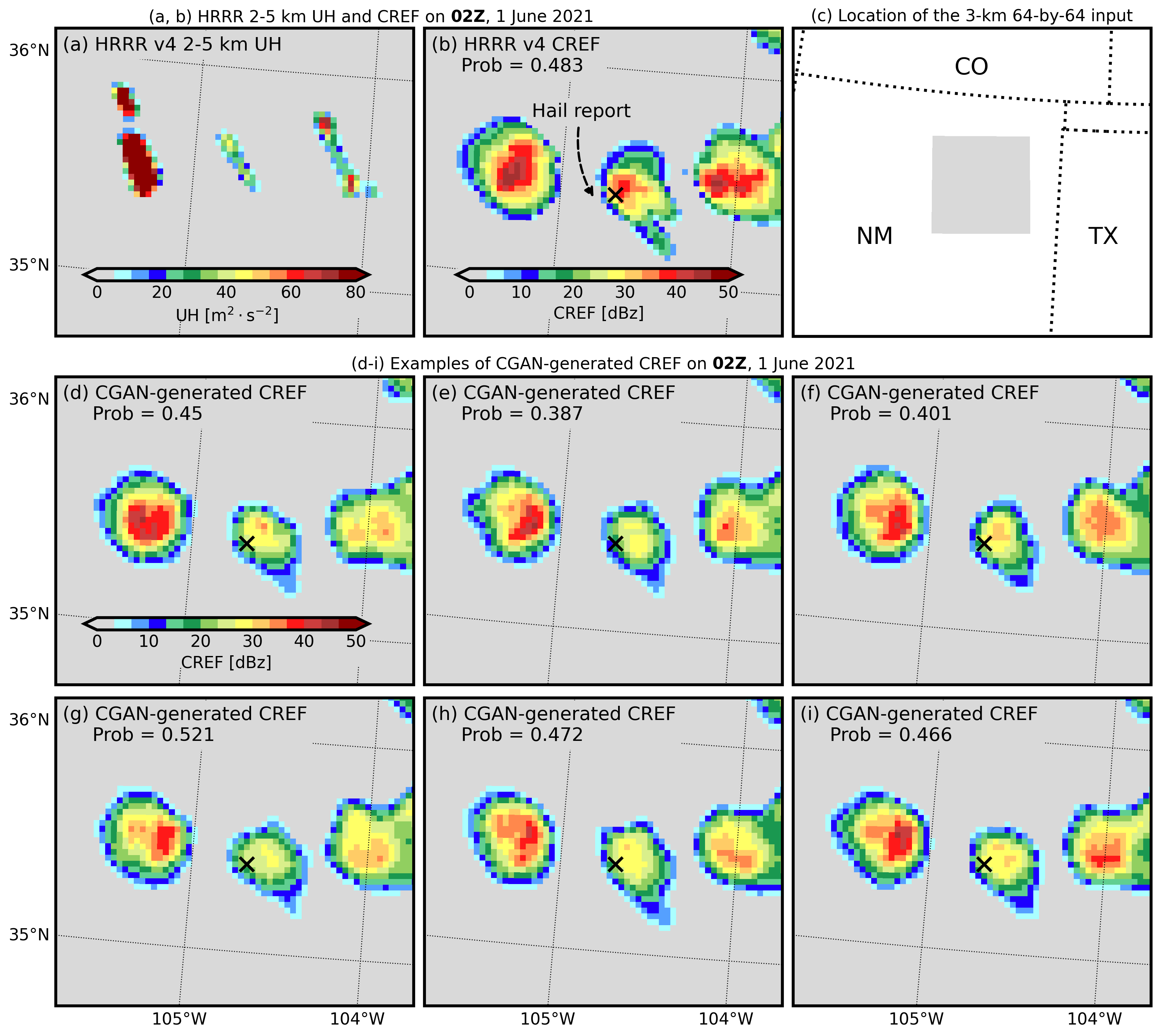}\\
 \caption{An example of severe weather prediction experiment on 1 June 2021 with 2-hr forecast lead time and 3-km grid spacing, 64-by-64 sized inputs. A hail event was reported (arrow and ``x'' marks) in this example. (a, b) 2-5 km UH and Max/Composite Radar Reflectivity produced by HRRRv4. (c) The geographical location of the input grid. (d-i) Synthetic Max/Composite Radar Reflectivity fields generated by CGAN; it takes (a) as one of the conditional inputs, and (b) with additive Gaussian noise as the initial state to generate (d-i).}\label{fig11}
\end{figure}

In this section, the CGAN-generated outputs are evaluated and compared to the original HRRRv4 forecasts. The purpose of this evaluation is to explain the good performance of the CGAN ensemble and identify desirable properties of generative models within the context of severe weather post-processing.

An example case on 0200 UTC 1 June 2021 is presented first. At this time, a set of single-cell thunderstorms were forecasted by the HRRRv4. A hail report occurred within the coverage of high CREF, indicating the HRRRv4 forecast was accurate. Part of the CGAN-generated CREF samples is shown in Fig.~\ref{fig11}d-i with the corresponding severe weather probabilities produced by the end-to-end CNN model. Comparing the CGAN outputs to the HRRRv4 forecasts, two performance highlights are evident:

\begin{enumerate}
    \item The CGAN-generated CREF patterns share roughly the same locations as their HRRRv4 counterparts (c.f. Fig.~\ref{fig11}b, d-i). This is because the sample-generation process of CGAN is constrained by conditional inputs such as 2-5 km UH. High-positive UH values point to the places of strong near-surface rotation and the CGAN is encouraged to generate CREF patterns at these locations (c.f. Fig.~\ref{fig11}.a and d-i). On the positive side, CGAN-generated samples would not exhibit large spatial discrepancies compared to the original HRRR forecasts and place negative impacts on severe weather prediction. In this example, given the ground truth of 1.0 (i.e., hail report) and generally correct HRRRv4 forecasts, the predicted severe weather probabilities were $\sim$0.4 for both CGAN-generated samples and the original HRRRv4 inputs. In some cases, when the HRRR forecasts contain large errors, the use of HRRR-originated conditional inputs may amplify these errors.

    \item Smaller scale differences can be found within the CGAN-generated samples. In Fig.~\ref{fig11}e, the intensity of the CREF pattern in the middle is weaker compared to the HRRRv4 sample, which results in a slightly lower severe weather probability, because the CNN model recognizes it as a weaker thunderstorm. In Fig.~\ref{fig11}g, the right-most and the middle CREF patterns are almost connected, with their zonal coverage being extended by the CGAN. The CNN model predicted a high probability on this sample, because the model may read it as two single cells being developed and merging. That said, although the CGAN-generated samples are similar to their HRRRv4 counterparts because of the shared conditional inputs, they still contain variations that can change the predicted probabilities of severe weather and create an ensemble with some useful uncertainty information.
\end{enumerate}

\begin{figure}[t]
 \noindent\includegraphics[width=\textwidth]{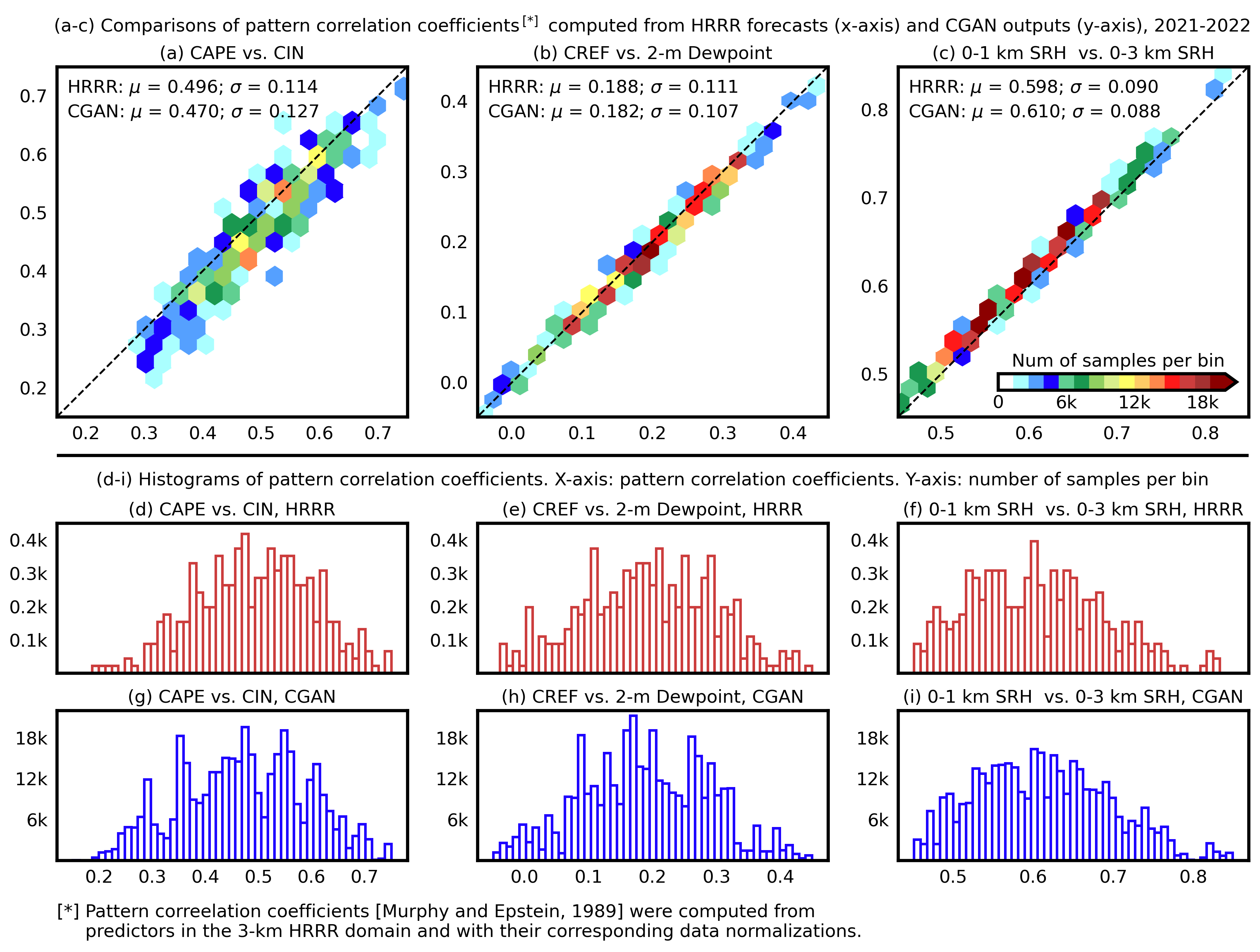}\\
 \caption{Comparisons of pattern correlation coefficients computed from HRRR and CGAN-generated predictors from 1 January 2021--31 December 2021 initializations and all forecast lead times. (a-c) 2-d histograms of correlation coefficients between CAPE and CIN, max/composite radar reflectivity and 2-m dew point temperature, and 0-1 km and 0-3 km SRH, respectively. The X-axis represents correlation coefficients computed from HRRR predictors. The Y-axis represents correlation coefficients computed from CGAN-generated predictors. The mean ($\mu$),  standard deviation ($\sigma$) of correlation coefficients, and diagonal reference lines are included. Colors represent the number of samples in each histogram bin. (d-f) 1-d histograms of correlation coefficients computed from HRRR predictors. (g-i) As in (d-f), but for CGAN-generated predictors.
}\label{fig12}
\end{figure}

Pattern correlations were computed between CAPE and CIN, CREF and 2-m Dewpoint, and 0-1 km SRH and 0-3 km SRH (Fig.~\ref{fig12}). These three pairs of predictors can be generated by the CGANs, and they exhibit the highest positive correlations among the combinations of the 15 predictors. The pattern correlations were computed for each initialization and forecast lead-time using all CONUS 3-km grid points.

We have two purposes for conducting this part of the evaluation. First, pattern correlations can show the physics-based consistency of CGAN outputs. For example, if the CGAN generates high CREF values, it is likely that it will generate high 2-m dewpoint values at the same location because strong hydrometer signals typically exist in a moist environment. Second, pattern correlations can reveal the impact of noise within the CGAN-generated samples. Given that the starting point of the sample-generation process is HRRR forecasts randomized with strong Gaussian noise [i.e., $\mathcal{N}(0, 0.5)$ compared to the normalized HRRR forecasts, which has standard deviations equal to or lower than 1.0; Section\ref{sec3}.\ref{sec31}], evaluating pattern correlations can make sure that the input noise does not damage the CGAN outputs.

Based on the 2-d correlation coefficient histograms, the CGAN-generated samples preserved the correlations that were present within the HRRRv4 forecasts (Fig.~\ref{fig12}). For CREF-to-2-m Dewpoint and 0-1 km SRH-to-0-3 km SRH, the distributions of correlation coefficients of the two fields were nearly identical (Fig.~\ref{fig12}b,c). For CAPE-to-CIN, the correlation coefficients of CGAN outputs exhibited a slightly larger variation, but quantitatively, they stayed within the range of [0.3, 0.7] (Fig.~\ref{fig12}a,d,g). These results also suggest that the impact of noise within CGAN outputs was minimal.

\begin{figure}[t]
 \noindent\includegraphics[width=\textwidth]{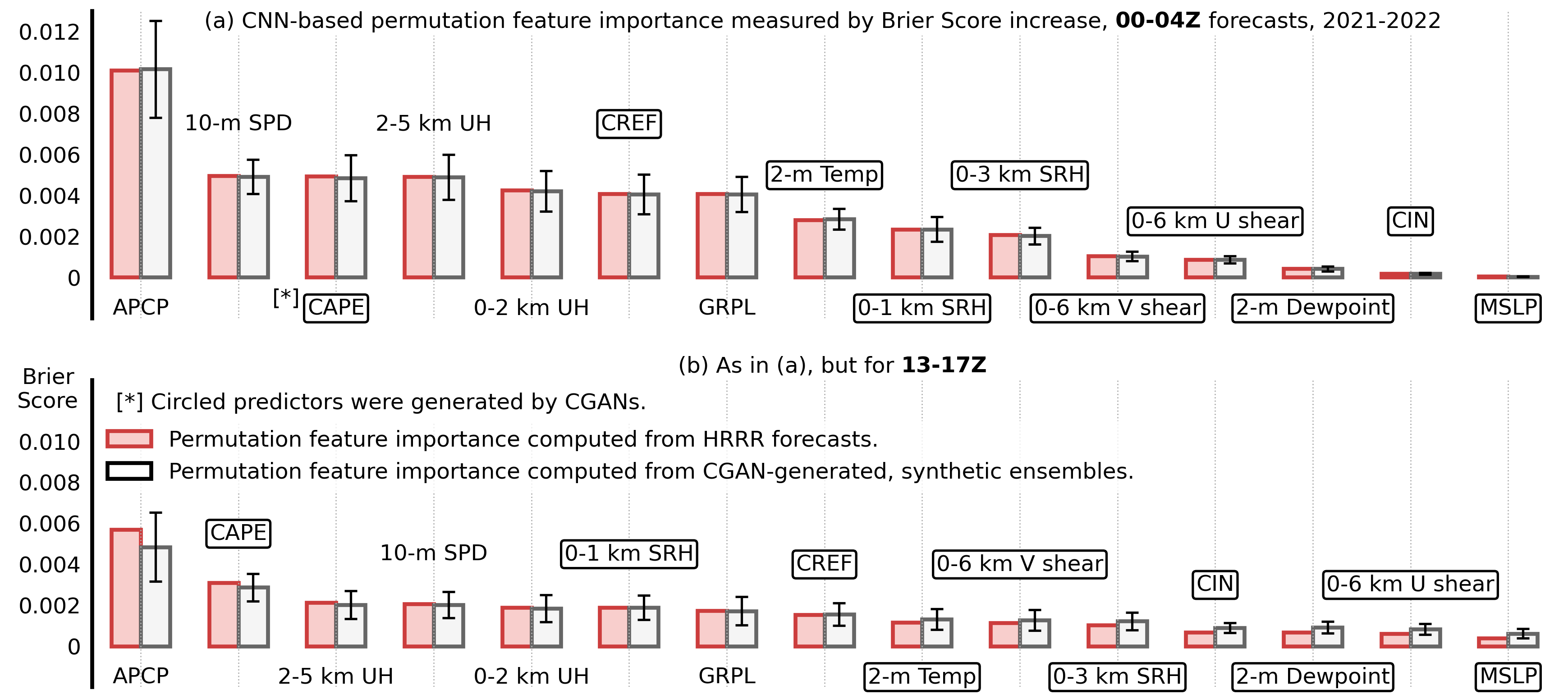}\\
 \caption{Comparisons of permutation feature importance of predictors from 1 January 2021--31 December 2021, measured by Brier Score increase (larger increase means more important) and the pre-trained CNN-based prediction model. (a) The permutation feature importance of 00-04Z forecasts. (b) As in (a) but for 13-17Z forecasts. Red- and white-colored bars represent the permutation feature importance of HRRR forecasts and CGAN-generated synthetic ensembles when using the pre-trained CNN-based prediction model, respectively. Error bars represent the permutation feature importance of individual synthetic ensemble members. For white-colored bars, circled predictors were generated by CGAN.}\label{fig13}
\end{figure}

Permutation feature importance measures the relative contribution of predictors in a given prediction system. That said, when the information of a predictor is permuted and loses its influence, a performance downgrade is expected. Here, the feature importance of CGAN outputs and HRRRv4 forecasts were calculated and compared. This evaluation is expected to examine how CGAN-generated ensembles would affect the decision-making of the CNN-based severe weather prediction model.

Storm-scale explicit predictors and CAPE are the most influential predictors, whereas 2-m dewpoint, MSLP, and CIN are the least influential predictors. The CGANs did not generate APCP, 10-m SPD, UH, and GRPL directly, however, these predictors still exhibit sufficiently large error bars or variations, in terms of their permutation importance per synthetic ensemble member. This indicates that, by generating CREF and environmental predictors, the CGANs have placed influences on the entire predictor set and the not-generated explicit predictors are affected as well.

For 00-04Z forecasts, predictors obtained from HRRRv4 and CGAN showed comparable permutation feature importance (Fig.~\ref{fig13}a). In other words, using HRRRv4 forecasts or CGAN outputs would not change its decision-making significantly. The feature importance of 13-17Z forecasts generally agrees with that of the 00-04Z forecasts. Taking the feature importance of HRRR forecasts as references, the CGAN outputs decreased the importance of high-influence predictors (e.g., CAPE) and increased the importance of low-influence predictors slightly (e.g., 2-m Dewpoint and MSLP; (Fig.~\ref{fig13}b)). This minor feature importance change is most likely explained by the increased HRRR forecast errors in longer forecast lead times. This forecast error impacts the quality of CCGAN outputs, and thus, decreases the usefulness of all predictors equally, resulting in slightly more uniform feature importance across predictors. Overall, the CGAN outputs were confirmed to play a positive role by expanding the ensemble without changing the importance of valuable predictors. 

\section{Discussion}\label{sec5}
This research applies CNNs and deep generative models to the ensemble post-processing of severe weather. Three research questions were raised in Section \ref{sec1}. The first question related to the performance of CNN-based severe prediction compared to simpler methods. Here, two CNN-based methods were verified and compared to an MLP baseline. Results indicate that the two CNN-based methods produced more skillful predictions with higher BSSs; they also have better resolution by predicting higher probabilities in those correctly verified severe weather cases; this outperformance is especially evident for 00-12Z forecasts. That said, CNN-based severe predictions are at least comparable to other state-of-the-art methods, and they are potentially better for short forecast lead times when the CAM inputs are skillful. An important technical choice that differs from prior works and leads to the success of CNNs is the decoupling of representation learning and classification. Severe weather prediction is a long-tailed classification problem with non-severe weather cases being the head class that accounts for more than 98\% of the samples. Without proper treatments, CNNs can overfit to the head class and ignore the tail class. The decoupling of representation learning and classification gives CNNs clearer learning goals in each of their inference stages. The majority of their hidden layers are trained to learn storm-scale representations from the data, whereas the classification layer can handle the sample imbalance by shifting its decision boundaries. We think CNNs, or other deep learning models that can process gridded inputs on an end-to-end basis, can potentially perform well for post-processing CAM output.

The second research question focused on the feasibility of deriving ensemble severe weather predictions from a deterministic CAM run using ML. In this research, MC dropout and CGAN-generated synthetic predictors were applied to produce ensembles of severe weather probabilities. The performance of these ensembles was examined in two ways. First, the BSSs of all methods were verified. Results indicated that ensemble severe weather predictions were beneficial because their ensemble means are more skillful than the deterministic individual members (Fig.~\ref{fig7}). Second, evaluations of uncertainty quantifications were conducted on the ensemble members. While there was good discard test performance, the spread-skill relationships indicated underdispersion. Although imperfect in terms of quantifying the predicted uncertainties, these ensembles, and especially the CGAN ensemble, are still practically useful, because their prediction errors are generally decreased with increasing discard rate. This means users can define thresholds of ensemble spreads based on discard rates and accept prediction results when ensemble spreads are below the thresholds only. Overall, the ensemble severe weather predictions of this research have met a certain level of success with improved forecast skills on ensemble mean and intuitive ensemble spreads that can distinguish good and bad predictions. 

The third research question addressed the contribution of CGANs. Based on the comparisons of CGAN ensembles and the CNN baseline in Section \ref{sec4}.\ref{sec42}, the use of CGAN outputs improved the ensemble mean BSS and its decompositions. In Section \ref{sec4}.\ref{sec43}, the CGAN ensembles have also shown better performances in spread-skill evaluations and discard tests. Thus, the CGANs have contributed positively in terms of severe weather prediction skill and uncertainty quantifications. In Section \ref{sec4}.\ref{sec44}, several desirable properties of CGANs have been identified: (1) The CGAN outputs can preserve the locations of convective cells while adding sufficient variations, e.g., the shape and central intensity of CREF patterns, that modifies the predicted severe weather probabilities (Fig.~\ref{fig11}). (2) The CGAN outputs preserved the inter-variable relationships as measured by pattern correlations (Fig.~\ref{fig12}) and were not impacted by the input random noise significantly. (3) CGAN-generated predictors have generally the same permutation feature importance as their original, HRRR-based counterparts (Fig.~\ref{fig13}). 

Future work could extend in several directions. First, the CGANs can participate in the training of the CNN-based severe weather prediction model as data augmentation. This idea was not implemented in this research because we intended to measure the contribution of CGANs in the inference stage—when the CGAN ensemble and the CNN baseline were operated on the same CNN model, their performance difference can be attributed to the use of CGAN outputs. Second, CGANs can be improved to generate CAM fields with larger variations. Here, two CGANs were applied to generate 10 predictors by taking 5 explicit storm-scale predictors, including 2-5 km UH, as conditional inputs. This configuration makes it easy for CGANs to generate convective cells in the same locations as their corresponding deterministic CAM forecasts, and it avoids the problem of creating non-existent storms under a stable environment. However, in situations where the CAM forecasts exhibit positional errors, the CGAN outputs may amplify that error. Developing better initial conditions for the sample generation process of CGAN may solve this problem. Finally, the purpose of the CGAN used here was to generate synthetic predictors from a deterministic CAM run. It offers a way to approximate the variability of the forecast data and is not tied to a specific post-processing model or region. For future research, other ML-based severe weather prediction models, e.g., dense neural networks and decision trees, can be applied to form an ensemble learning system. The method that combines CGANs and severe weather post-processing models can also potentially be generalized to other regions where high-quality severe weather observations are available. 

\section{Conclusions}\label{sec6}

A novel post-processing method was proposed by incorporating Conditional Generative Adversarial Networks (CGANs) and a Convolutional Neural Network (CNN) classifier to generate probabilistic forecasts of convective weather hazards from deterministic convection-allowing model (CAM) forecasts. The CGANs were trained to create synthetic predictors from the deterministic CAM fields, whereas the CNN classifier takes the CGAN outputs as inputs and produces severe weather probabilities on an end-to-end basis. Monte Carlo (MC) dropout was also implemented within the CNN classifier for the purpose of ensemble learning and uncertainty quantification.

The method was tested with the High-Resolution Rapid Refresh (HRRR) version 4 forecasts and verified using severe weather reports collected by the Storm Prediction Center (SPC reports) over the Conterminous United States (CONUS) from 1 January 2021--31 December 2021. Based on the verification results, our method produced better severe weather predictions compared to baselines that were trained without CGAN outputs. For the prediction skill of the ensemble mean, our method reached 0.2 Brier Skill Score (BSS) for short forecast lead times, and stayed around 0.1 BSS for longer forecast lead times, showing up to 20\% BSS increase compared to the baselines. Spatial analysis indicates that the BSS increase is primarily contributed from the northeastern United States and the Great Plains where severe weather is observed frequently. Comparisons of Brier Score components showed that the CGAN ensembles had better resolution by issuing higher probabilities on verified severe weather cases. For the evaluation of uncertainty quantification, the CGAN ensemble produced better spread-skill reliability compared to the baselines. The CGAN ensemble also performed the best in discard tests, indicating that its ensemble spread distinguished good and bad predictions. Finally, the CGAN outputs were evaluated with examples, pattern correlations, and permutation feature importance. Results indicated that the CGANs preserved convective-scale information at the same locations as their HRRR counterpart, meanwhile adding sufficient variations. The inter-variable correlations and the predictability of influential predictors were also similar between CGAN outputs and HRRR forecasts.

To our knowledge, no previous research has experimented with the combination of deep generative models and CNNs within the context of severe weather post-processing. Using CGANs may bridge the gap between deterministic CAM forecasts and the ensemble prediction of severe weather. More broadly, it also provides a formulation of how to design, implement, and verify deep generative models for solving weather forecasting challenges.

\clearpage
\acknowledgments
The authors thank Dr. Terra Ladwig, Global Systems Laboratory, NOAA, for making the experimental HRRR dataset available for use in this work. This material is based upon work supported by the National Center for Atmospheric Research (NCAR), which is a major facility sponsored by the National Science Foundation (NSF) under Cooperative Agreement No. 1852977. This research was supported by NOAA OAR grant NA19OAR4590128, the NSF NCAR Short-term Explicit Prediction Program, and NSF Grant No. ICER-2019758. Supercomputing support was provided by NSF NCAR Cheyenne and Casper (Computational and Information Systems Laboratory, CISL 2020). The authors also thank Dr. Craig Schwartz and David Ahijevych at NSF NCAR and three anonymous reviewers for their feedback.

%
%
\datastatement
Our data pre-processing, training, and data visualization code can be found at \url{https://github.com/yingkaisha/AIES_D_23_0094}. A frozen release of the source code is available at \url{https://doi.org/10.5281/zenodo.10732687}. The HRRRv3 and HRRRv4 forecasts are available at \url{https://storage.googleapis.com/high-resolution-rapid-refresh/}. Our pre-processed data and the HRRRv4 reforecasts are available upon request.


%






%



\bibliographystyle{ametsocV6}
\bibliography{references}

\end{document}